\documentclass{article}

 \usepackage[preprint]{neurips_2026}

\usepackage[utf8]{inputenc} 
\usepackage[T1]{fontenc}    
\usepackage{hyperref}       
\usepackage{url}            
\usepackage{booktabs}       
\usepackage{amsfonts}       
\usepackage{nicefrac}       
\usepackage{microtype}      
\usepackage{xcolor}         
\usepackage{multirow} 

\usepackage{graphicx}
\usepackage{placeins}
\usepackage{xcolor}
\usepackage[table]{xcolor}
\usepackage{float}

\definecolor{mfcol}{RGB}{255,221,190}   
\definecolor{imfcol}{RGB}{204,242,210}  
\definecolor{soflowcol}{RGB}{208,224,255} 
\definecolor{raecol}{RGB}{231,214,255}  
\definecolor{scaleraecol}{RGB}{255,245,191} 
\definecolor{sdcol}{RGB}{200,235,245}   
\definecolor{fluxcol}{RGB}{255,206,230} 
\definecolor{sitcol}{RGB}{223,237,207}  

\newcommand{\bestblue}[1]{\textcolor{blue}{\textbf{#1}}}

\usepackage[compact]{titlesec}

\titlespacing{\section}{0pt}{0.2ex plus 0.1ex minus 0.1ex}{0.2ex}
\titlespacing{\subsection}{0pt}{0.1ex plus 0.1ex minus 0.1ex}{0.1ex}
\usepackage{siunitx}
\newtheorem{lemma}{Lemma}

\usepackage{siunitx}
\title{Setting-Matched and Semantics-Scaled Benchmarking of One-Step Generative Models Against Multistep Diffusion and Flow Models}

\begin{document}
\author{%
    \textbf{Advaith Ravishankar}$^{1,*}$ \quad
    \textbf{Serena Liu}$^{1,*}$ \quad
    \textbf{Mingyang Wang}$^{1,*}$ \quad
    \textbf{Todd Zhou}$^{1}$ \\[1pt]
    \textbf{Jeffrey Zhou}$^{1}$ \quad
    \textbf{Arnav Sharma}$^{1}$ \quad
    \textbf{Ziling Hu}$^{1}$ \quad
    \textbf{Léopold Das}$^{1}$ \\[1pt]
    \textbf{Abdulaziz Sobirov}$^{1}$ \quad
    \textbf{Faizaan Siddique}$^{1}$ \quad
    \textbf{Freddy Yu}$^{1}$ \quad
    \textbf{Seungjoo Baek}$^{1}$ \\[1pt]
    \textbf{Yan Luo}$^{1,\dagger}$ \quad
    \textbf{Mengyu Wang}$^{1,\dagger}$ \\[5pt]
    $^{1}$Harvard AI and Robotics Lab, Harvard University \\[2pt]
    $^{*}$Equal contribution \quad
    $^{\dagger}$Equal contribution as co-senior authors
}

\maketitle

\begin{abstract}
State-of-the-art text-to-image models produce high-quality images, but inference remains expensive as generation requires several sequential ODE or denoising steps. Native one-step models aim to reduce this cost by mapping noise to an image in a single step, yet fair comparisons to multi-step systems are difficult because studies use mismatched sampling steps and different classifier-free guidance (CFG) settings, where CFG can shift FID, Inception Score, and CLIP-based alignment in opposing directions. It is also unclear how well one-step models scale to multi-step inference, and there is limited standardized out-of-distribution evaluation for label-ID-conditioned generators beyond ImageNet. To address this, we benchmark eight models spanning one-step flows (MeanFlow, Improved MeanFlow, SoFlow), multi-step baselines (RAE, Scale-RAE), and established systems (SiT, Stable Diffusion 3.5, FLUX.1) under a class-conditional protocol on ImageNet validation, ImageNetV2, and reLAIONet, our new proofread out-of-distribution dataset aligned to ImageNet label IDs. Using FID, Inception Score, CLIP Score, and Pick Score, we show that FID-focused model development and CFG selection can be misleading in few-step regimes, where guidance changes can improve FID while degrading text-image alignment and human preference signals, worsening visual quality. To make these tradeoffs explicit, we introduce CLIP-scaled and PickScore-scaled variants of FID (csFID, psFID) and Inception Score (csIS, psIS) to serve as a diagnostic for semantically aligned image generation. The code for reproducing the benchmarks is made publicly available at \href{https://github.com/Harvard-AI-and-Robotics-Lab/FairBenchmarkingFlow}{github.com/Harvard-AI-and-Robotics-Lab/FairBenchmarkingFlow}
\end{abstract}

\section{Introduction}
Recent progress in image generation has been driven by diffusion and flow-based models that produce realistic samples, but require several sequential denoising or ODE steps at inference. This makes state-of-the-art systems expensive to deploy, since generating a single image can require dozens of forward passes through billion-parameter backbones. Motivated by this bottleneck, recent work has targeted native one-step generation, where a single model evaluation maps noise directly to an image. While early efforts relied on distillation, newer methods now report strong one-step results without a teacher, including MeanFlow~\cite{geng2025meanflow}, Improved MeanFlow (iMF)~\cite{geng2025improvedmeanflow}, and SoFlow~\cite{luo2025soflow}. Yet, comparing one-step models against established multi-step systems remains nontrivial. First, models are often reported under heterogeneous inference settings, such as label-ID conditioning versus text conditioning and different classifier-free guidance (CFG) values, which alter metrics and appearance. Second, evaluation is often centered around Fr\'echet Inception Distance (FID), which does not directly capture text-image alignment or human preference. These factors make it unclear how competitive one-step models are under matched guidance and step settings, and whether optimizing FID alone yields the best visual quality.

In this work, we present a setting-matched and semantics-scaled benchmark of emerging one-step flow models against established diffusion and flow-based systems under consistent inference protocols. We evaluate eight approaches that span native one-step models (MeanFlow, iMF, SoFlow), flow baselines (Scalable Interpolant Transformer (SiT)~\cite{ma2024sit}, Representation Autoencoder (RAE)~\cite{zheng2025ditrae}, Scale-RAE~\cite{tong2026scalerrae}), and state-of-the-art text-to-image systems (Stable Diffusion 3.5 Large~\cite{stabilityai_sd3_techreport}, FLUX.1~\cite{blackforestlabs_flux_modelcard}). For each method, we compare matched guidance (CFG of 7 where applicable) and each model's native guidance. We also evaluate both one-step and multi-step inference (1 and 25 steps). To test generalization beyond ImageNet, we introduce \textbf{reLAIONet}, a manually proofread evaluation dataset mapping web-sourced images to ImageNet classes that provides a larger, more diverse out-of-distribution benchmark while preserving label-ID conditioning.

\begin{figure}[!t]
    \centering
    \makebox[\linewidth][c]{%
    \includegraphics[width=1.3\textwidth]{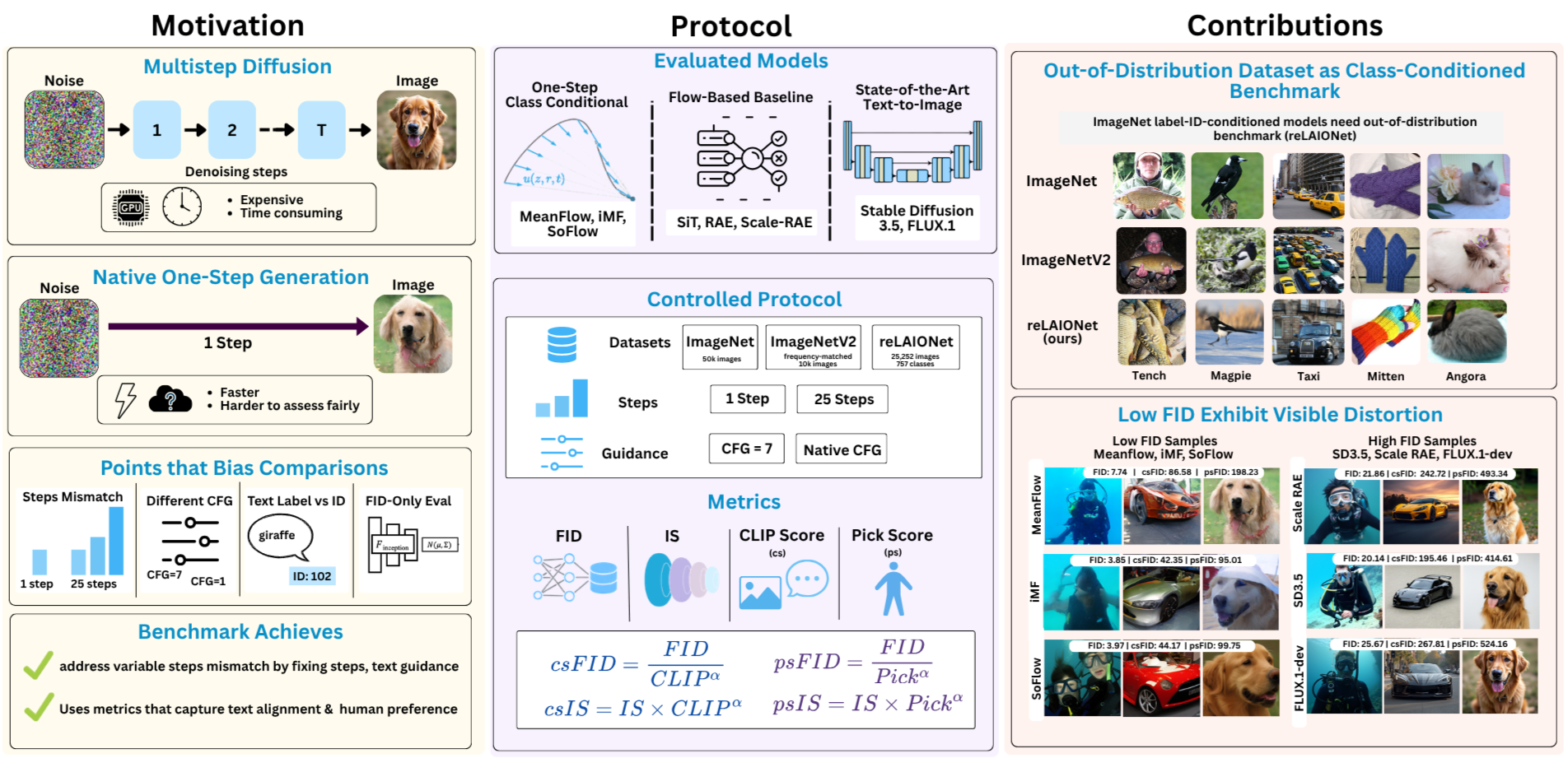}
    }
    \caption{\textbf{Benchmark overview.}
Left: Motivation for fair one-step vs. multi-step comparison. Middle: setting-matched and semantics-scaled protocol over models, datasets, steps, CFG, and metrics. Right: reLAIONet preserves the ImageNet label-ID interface under distribution shift, while qualitative examples show that low FID hides visual artifacts, motivating csFID, psFID, csIS, and psIS.}
    \label{fig:qualitative_results}
\end{figure}

Finally, we study how metric choice shapes conclusions about generative quality. We find that optimizing FID alone can improve distributional similarity but degrade Inception Score, CLIP Score, Pick Score, and perceptual quality. To expose tradeoffs in semantic-alignment and human preference, we introduce CLIP- and Pick-Score-scaled variants of FID and IS: csFID, psFID, csIS, and psIS. These metrics show that the strongest native one-step models narrow the quantitative gap to SD3.5 Large and FLUX.1 under multi-step class-conditional inference, though qualitative distortions remain. Overall, our results make three points: (1) fair comparison of one-step and multi-step generators requires matched step and guidance settings, since step count and CFG shift different metrics in opposing directions. (2) label-ID-conditioned models need out-of-distribution benchmarks such as reLAIONet that preserve the ImageNet interface to distinguish generalizable generation quality from distributional overfitting. (3) FID-only evaluation hides image generation quality, which our semantics-scaled metrics (csFID, psFID, csIS, and psIS) make explicit.

\section{Related Work}
\subsection{Diffusion and Flow Models}
Since 2020, diffusion-based approaches have dominated visual generation. DDPMs introduced iterative denoising through sequential stochastic steps~\cite{ho2020ddpm}, while latent diffusion and deterministic ODE-style sampling improved efficiency and reduced timestep requirements~\cite{rombach2022ldm,song2021ddim}. This motivated flow-based formulations. Flow Matching regressed vector fields along probability paths~\cite{lipman2023flow}, Rectified Flow encouraged straighter trajectories~\cite{liu2023flow}, and SiT unified diffusion and flow objectives within a scalable interpolant framework~\cite{ma2024sit}. However, many state-of-the-art text-to-image systems that adopted flow-style designs~\cite{stabilityai_sd3_techreport,blackforestlabs_flux_modelcard} are still computationally expensive. To address this, recent work targeted one-step generation. Rectified Flow and InstaFlow showed that distillation can enable few- and one-step generation~\cite{liu2024instaflow}. Native one-step models trained without distillation also emerged. MeanFlow learned an average velocity field for one-step ImageNet generation~\cite{geng2025meanflow}. iMF reformulated the objective as instantaneous velocity regression~\cite{geng2025improvedmeanflow}. SoFlow combined Flow Matching with a solution-consistency objective~\cite{luo2025soflow}, and Drifting Models later framed training as learning a drift field that progressively transforms generated distributions toward the image distribution~\cite{deng2026drifting}.

\begin{figure}[!t]
    \centering
    \makebox[\linewidth][c]{%
    \includegraphics[width=1\textwidth]{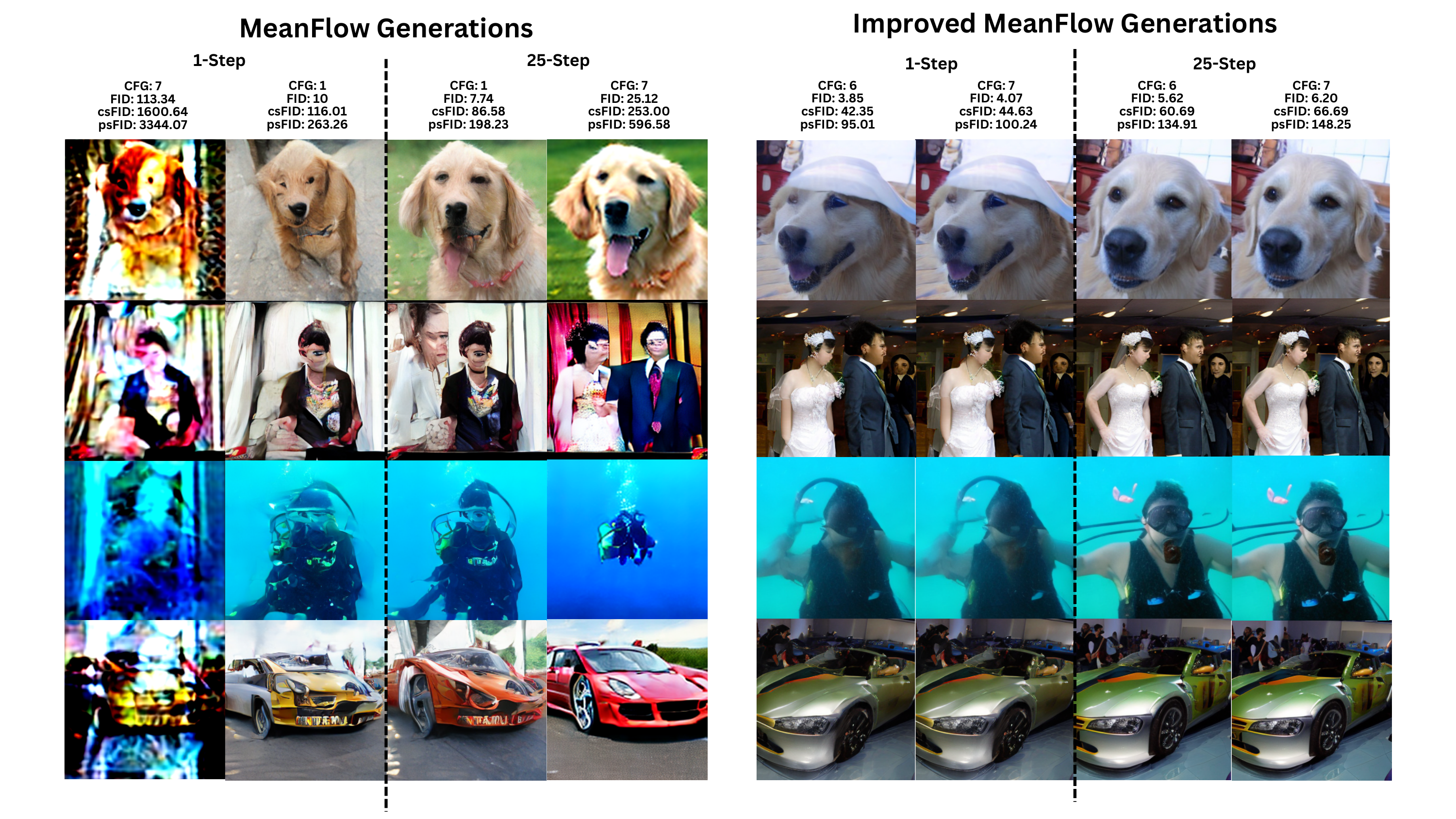}
    }
    \caption{Within model family comparison for native-one step models. Images generated with classes Golden Retriever, GoldFish, Groom, Scuba Diver and Sports Car for Meanflow and iMF. Both iMF and Meanflow show clear distortions even with low FID, indicating that FID minimization alone does not imply better image generation}
    \label{fig:within_qualitative_results}
\end{figure}

\subsection{Challenges with Benchmarking Protocol, Evaluation Metrics, and Datasets}

State-of-the-art text-to-image systems such as Stable Diffusion 3.5 are trained on large captioned corpora with text conditioning and multi-step CFG sampling~\cite{ho2022cfg,stabilityai_sd3_techreport}, while one-step flow models such as MeanFlow, iMF, and SoFlow are trained on ImageNet with label-ID conditioning and reported under one- or few-step settings~\cite{geng2025meanflow,geng2025improvedmeanflow,luo2025soflow,deng2026drifting}. Different approaches also use different CFG scales, often tuned to optimize FID. These mismatches in conditioning type, step count, and guidance make comparisons nontrivial, motivating controlled evaluation under matched settings. Moreover, it remains unclear how additional inference steps affect one-step model performance.

Image generation evaluation has centered around Fr\'echet Inception Distance (FID) and Inception Score (IS)~\cite{heusel2017fid,salimans2016is}, which primarily measure similarity in InceptionNet feature space rather than human-perceived realism. As models improved, alignment- and preference-oriented metrics such as CLIP Score and PickScore emerged to better capture semantic consistency and perceived quality~\cite{radford2021clip,kirstain2023pickscore}. However, inference heuristics such as CFG modulate each metric differently, so reported results often rely on empirically chosen CFG settings tuned for FID. This further complicates comparisons between multi-step diffusion systems and one-step flow models.

Established text-to-image generators are trained on text-image corpora such as LAION, enabling open-vocabulary prompt-based evaluation~\cite{schuhmann2022laion}. This allows generalization evaluation with any image with a caption. However, native one-step models are trained and validated on ImageNet using label IDs~\cite{geng2025meanflow,geng2025improvedmeanflow}, requiring evaluation datasets that preserve the same class labels. ImageNetV2 provides such an external test set, but is relatively small (10,000 images). This motivates an additional dataset to measure generalization for ImageNet-trained generators.

\begin{figure}[!t]
    \centering
    \makebox[\linewidth][c]{%
    \includegraphics[width=1.2\textwidth]{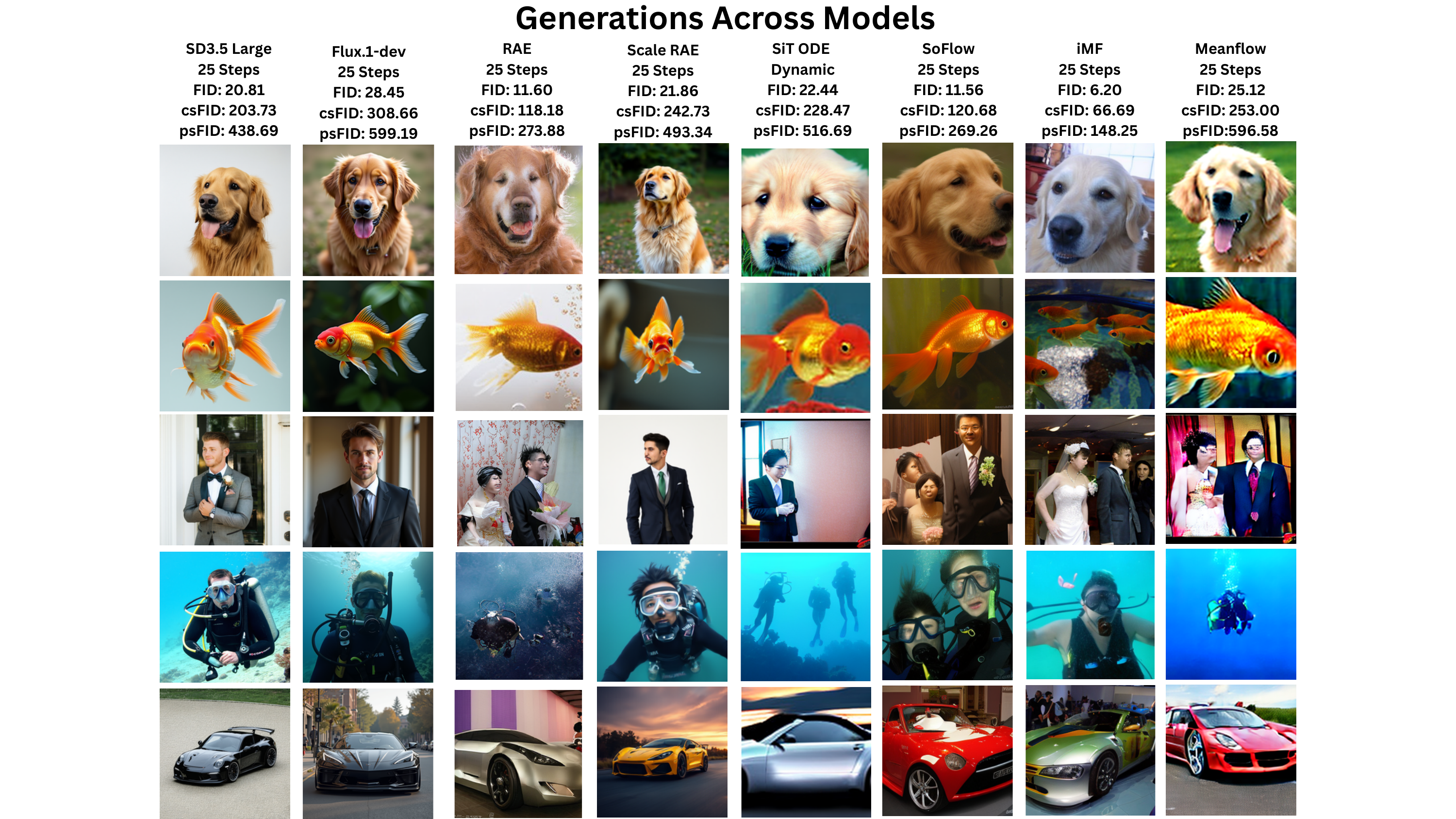}
    }
    \caption{Samples for Golden Retriever, GoldFish, Groom, Scuba Diver, and Sports Car at 25 steps and CFG 7. One-step models show low-FID models but exhibit artifacts in faces, underwater scenes, and fine details, while higher-FID SOTA models (SD3.5 and Flux) produce coherent samples. This motivates reporting scaled metrics such as csFID, psFID, csIS, and psIS alongside FID.}
    \label{fig:qualitative_results}
\end{figure}

\section{Benchmark}
\subsection{Benchmarking Setup}
Our benchmark evaluates whether one-step generators are competitive with multi-step systems under matched CFG and step settings, whether they improve with step scaling, and whether csFID, psFID, csIS, and psIS better diagnose generation quality. We compare MeanFlow, iMF, and SoFlow against SiT, RAE, Scale-RAE, SD3.5, and FLUX.1 using the largest public checkpoints. We evaluate CFG 7 and each model's reported CFG at 1 and 25 steps on ImageNet validation set, ImageNetV2, and reLAIONet.

\subsection{Design Choices} \label{sec:design_choices} 
\subsubsection{CFG Choices} Higher CFG makes outputs better match the prompt, but can make them less diverse and more artifact-prone. On the other hand, lower CFG means weaker text guidance, ignoring prompt details. To find the correct balance, we set CFG to 7 to match common practice in SD-based pipelines and prior SD-based editing work~\cite{brooks2022instructpix2pix}. In addition, we also use the reported CFG setting of each model as it provides single-step and multi-step performance comparison, allowing us to understand the changes between fixed and tuned guidance scales.

\subsubsection{Step Choices and NFE}
We evaluate all models in both a one-step and a multi-step regime. Following common Flow Matching grids (NFE $\in \{1,2,4,8,16,32\}$), we use 25 steps as a strong multi-step anchor alongside 1 step. For consistency, we report ODE/denoising steps rather than number of function evaluations (NFE), since NFE per step varies by model: iMF and SoFlow collapse conditional and unconditional velocities, enabling 1 NFE per step, whereas SD3.5, RAE, Scale-RAE, FLUX.1, and SiT require at least 2 NFEs per step. Although MeanFlow reports 1 NFE per step, this setting ignores the conditioned velocity (i.e., removes guidance). To maintain a guided generation protocol, we instead use the training-time formulation, which requires 2 NFEs per step. Given these model-specific NFE differences, reporting step count provides the clearest cross-model comparison.

\subsubsection{Implementation Details} \label{sec:implementation_details}
SoFlow bakes CFG into its weights, so only step count can be varied. SiT uses the adaptive \texttt{dopri5} solver, so we keep the reported solver rather than replacing it with Euler for explicit step control. MeanFlow, iMF, SoFlow, and RAE are ImageNet label-ID-conditioned, while SD3.5, FLUX.1-dev, and Scale-RAE are text-conditioned. We prompt SD3.5 and FLUX.1-dev with ``a photo of a \{class\_name\}'' and Scale-RAE with ``Can you generate a photo of a \{class\_name\}?'' due to its Qwen2 conversational interface. This comparison contextualizes one-step models against strong generators, but disadvantages text-conditioned systems under the ImageNet label-ID protocol.


\subsection{Metrics}
We report Fr\'echet Inception Distance (FID), Inception Score (IS), CLIP Score, and PickScore. FID and IS measure distributional quality, classifiability, and diversity. CLIP Score measures semantic alignment and PickScore estimates human preference. For each dataset, we compute FID against the full reference set using ordered generation from the dataset classes. For CLIP Score and PickScore, each image is evaluated against ``a photo of a \{class\_name\}''. Since one-step models are often tuned toward FID, we report CLIP-scaled and PickScore-scaled variants of FID and IS:
\begin{equation}
\mathrm{csFID}_{\alpha}=\frac{\mathrm{FID}}{\mathrm{CLIP}^{\alpha}},\quad
\mathrm{psFID}_{\alpha}=\frac{\mathrm{FID}}{\mathrm{Pick}^{\alpha}},\quad
\mathrm{csIS}_{\alpha}=\mathrm{IS}\cdot \mathrm{CLIP}^{\alpha},\quad
\mathrm{psIS}_{\alpha}=\mathrm{IS}\cdot \mathrm{Pick}^{\alpha}.
\end{equation}
We use raw CLIP Score and PickScore values in $[0,1]$ as alignment and preference weights. These scores vary in a narrow range across methods, so $\alpha$ controls how strongly alignment and preference differences are amplified. We set $\alpha=2$ in the main paper because it stretches the compressed CLIP and PickScore ranges while preserving stable rankings. We report correlations between the semantics-scaled metrics with base metrics and $\alpha \in \{1,2\}$ in the appendix. Each metric captures the following: (1) csFID measures distributional similarity to real images weighted by semantic alignment, (2) psFID measures distributional similarity weighted by human preference, (3) csIS measures class separability and diversity weighted by semantic alignment, and (4) psIS measures class separability and diversity weighted by human preference. Lower csFID and psFID and higher csIS and psIS indicate stronger overall generation quality.

\subsection{Datasets}
\subsubsection{ImageNet Family}
The one-step generation models are trained on ImageNet~\cite{deng2009imagenet} (ILSVRC 2012), which contains 1.28M training images and 50{,}000 validation images across 1{,}000 object categories. We use the full validation split as the in-domain reference distribution and generate one image per validation label, yielding 50{,}000 generated samples. We also evaluate on ImageNetV2~\cite{recht2019imagenetv2}, a frequency-matched test set that uses the same 1{,}000-class label with 10{,}000 images. We use it as a within-domain generalization test, where drops from ImageNet validation to ImageNetV2 indicate sensitivity to the original validation distribution.

\subsubsection{reLAIONet}
Since the one-step models in this benchmark are trained on ImageNet, evaluation is limited to ImageNet validation and ImageNetV2~\cite{recht2019imagenetv2}, which share the same label space and similar photographic distribution as the training data. To test broader generalization while preserving ImageNet label-ID conditioning, we introduce reLAIONet, a proofread web-sourced evaluation set built from LAIONet~\cite{shirali2023imagenetlaion}. Since many original LAION-400M URLs are unavailable, we constructed reLAIONet from reLAION-400M~\cite{schuhmann2022laion}. Applying the pipeline yields images for up to 997 ImageNet classes. We download up to 70 images per class and manually annotate them with 12 reviewers through two rounds of filtering: the first round removes clear mismatches, and the filtered set is passed to a second round of reviewers for additional verification, with both rounds assigning reviewers to different image subsets. This removes mislabeled and ambiguous samples, producing 25,252 images across 757 ImageNet classes. The final class distribution is reported in the appendix. Since reLAIONet images come from open web crawls rather than Flickr, the primary source of ImageNet, they introduce different photographic styles and contexts while retaining the label-ID interface.

\begin{table}[!t]
\centering
\small
\setlength{\tabcolsep}{1.8pt}
\renewcommand{\arraystretch}{1.1}
\makebox[\linewidth][c]{%
\begin{tabular}{lcccccccccc}
\toprule
\textbf{Model} &
\textbf{Steps} & \textbf{CFG} &
\textbf{FID $\downarrow$} & \textbf{IS $\uparrow$} &
\textbf{CLIP $\uparrow$} & \textbf{Pick $\uparrow$} &
\textbf{csFID$_2$ $\downarrow$} & \textbf{psFID$_2$ $\downarrow$} &
\textbf{csIS$_2$ $\uparrow$} & \textbf{psIS$_2$ $\uparrow$} \\
\midrule

Meanflow (B/4) & 1 & 7.0 & 113.34 & 13.93 & 26.61 & 18.41 & 1600.64 & 3344.07 & 0.99 & 0.47 \\
\textbf{Meanflow (B/4)} & 25 & 7.0 & 25.12 & \textbf{346.19} & \textbf{31.51} & \textbf{20.52} & 253.00 & 596.58 & \textbf{34.37} & \textbf{14.58} \\
Meanflow (B/4)$^{\star}$ & 1 & 1.0 & 10.00 & 150.47 & 29.36 & 19.49 & 116.01 & 263.26 & 12.97 & 5.72 \\
Meanflow (B/4)$^{\star}$ & 25 & 1.0 & \textbf{7.74} & 205.27 & 29.90 & 19.76 & \textbf{86.58} & \textbf{198.23} & 18.35 & 8.02 \\
\midrule

iMF (XL/2) & 1 & 7.0 & 4.07 & 293.18 & 30.20 & 20.15 & 44.63 & 100.24 & 26.74 & 11.90 \\
\textbf{iMF (XL/2)} & 25 & 7.0 & 6.20 & \textbf{346.87} & \textbf{30.49} & \textbf{20.45} & 66.69 & 148.25 & \textbf{32.25} & \textbf{14.51} \\
iMF (XL/2)$^{\star}$ & 1 & 6.0 & \textbf{3.85} & 285.50 & 30.15 & 20.13 & \textbf{42.35} & \textbf{95.01} & 25.95 & 11.57 \\
iMF (XL/2)$^{\star}$ & 25 & 6.0 & 5.62 & 336.84 & 30.43 & 20.41 & 60.69 & 134.91 & 31.19 & 14.03 \\
\midrule

SoFlow (XL/2)$^{\star}$ & 1 & 2.0 & \textbf{3.97} & 238.98 & 29.98 & 19.95 & \textbf{44.17} & \textbf{99.75} & 21.48 & 9.51 \\
\textbf{SoFlow (XL/2)$^{\star}$} & 25 & 2.0 & 11.56 & \textbf{362.88} & \textbf{30.95} & \textbf{20.72} & 120.68 & 269.26 & \textbf{34.76} & \textbf{15.58} \\
\midrule

RAE & 1 & 7.0 & 175.00 & 6.69 & 24.98 & 18.88 & 2804.49 & 4909.46 & 0.42 & 0.24 \\
\textbf{RAE} & 25 & 7.0 & 11.60 & \textbf{382.36} & \textbf{31.33} & \textbf{20.58} & 118.18 & 273.88 & \textbf{37.53} & \textbf{16.19} \\
RAE$^{\star}$ & 1 & 1.42 & 302.76 & 3.43 & 24.08 & 18.71 & 5221.38 & 8648.70 & 0.20 & 0.12 \\
RAE$^{\star}$ & 25 & 1.42 & \bestblue{2.61} & 255.53 & 30.31 & 20.43 & \bestblue{28.41} & \bestblue{62.53} & 23.48 & 10.67 \\
\midrule

Scale RAE & 1 & 7.0 & 317.30 & 1.53 & 20.39 & 16.89 & 7631.95 & 11122.71 & 0.06 & 0.04 \\
Scale RAE & 25 & 7.0 & \textbf{21.86} & \textbf{90.49} & 30.01 & 21.05 & \textbf{242.73} & \textbf{493.34} & 8.15 & \textbf{4.01} \\
Scale RAE$^{\star}$ & 1 & 1.42 & 317.55 & 1.53 & 20.39 & 16.89 & 7637.97 & 11131.48 & 0.06 & 0.04 \\
\textbf{Scale RAE$^{\star}$} & 25 & 1.42 & 21.96 & 90.40 & \textbf{30.03} & \textbf{21.06} & 243.51 & 495.13 & \textbf{8.15} & 4.01 \\
\midrule

\textbf{SiT ODE} & Dynamic & 7.0 & 22.44 & \bestblue{466.54} & \textbf{31.34} & \textbf{20.84} & 228.47 & 516.69 & \bestblue{45.82} & \bestblue{20.26} \\
SiT ODE$^{\star}$ & Dynamic & 1.5 & \textbf{3.83} & 254.45 & 30.00 & 20.17 & \textbf{42.56} & \textbf{94.14} & 22.90 & 10.35 \\
\midrule

SD3.5 Large & 1 & 7.0 & 179.07 & 2.71 & 24.42 & 18.31 & 3002.84 & 5341.29 & 0.16 & 0.09 \\
SD3.5 Large & 25 & 7.0 & 20.81 & 202.15 & 31.96 & 21.78 & 203.73 & 438.69 & 20.65 & 9.59 \\
SD3.5 Large$^{\star}$ & 1 & 3.5 & 186.03 & 2.68 & 24.52 & 18.60 & 3094.16 & 5377.21 & 0.16 & 0.09 \\
\textbf{SD3.5 Large$^{\star}$} & 25 & 3.5 & \textbf{20.14} & \textbf{214.58} & \bestblue{32.10} & \textbf{22.04} & \textbf{195.46} & \textbf{414.61} & \textbf{22.11} & \textbf{10.42} \\
\midrule

Flux.1-dev & 1 & 7.0 & 183.09 & 3.97 & 24.49 & 18.90 & 3052.72 & 5125.56 & 0.24 & 0.14 \\
Flux.1-dev & 25 & 7.0 & 28.45 & 115.62 & 30.36 & 21.79 & 308.66 & 599.19 & 10.66 & 5.49 \\
Flux.1-dev$^{\star}$ & 1 & 3.5 & 196.95 & 3.89 & 25.01 & 19.10 & 3148.68 & 5398.70 & 0.24 & 0.14 \\
\textbf{Flux.1-dev$^{\star}$} & 25 & 3.5 & \textbf{25.67} & \textbf{148.98} & \textbf{30.96} & \bestblue{22.13} & \textbf{267.81} & \textbf{524.16} & \textbf{14.28} & \textbf{7.30} \\

\bottomrule
\end{tabular}
}
\caption{ImageNet validation benchmark. $^{\star}$ denotes model-specific tuned CFG. SoFlow prevents custom CFG and SiT prevents explicit step control; see \ref{sec:implementation_details}. We report FID, IS, CLIP, Pick, and semantics-scaled metrics csFID$_2$, psFID$_2$, csIS$_2$, and psIS$_2$. Bold marks the best within each model family. Blue marks the best value across all models for each metric.}
\label{tab:image_benchmark_performance}
\end{table}

\section{Results}



\subsection{Impact of Sampling Steps}
As seen in Table~\ref{tab:image_benchmark_performance}, the FID numbers increased from one-step to 25-step sampling for the one-step generation models including iMF (from 3.85 to 5.62) and SoFlow (from 3.97 to 11.56), yet the FID number decreased for the one-step generation model MeanFlow (from 10.00 to 7.74). The FID numbers decreased from one-step to 25-step sampling for the multistep generation models including RAE (from 317.55 to 2.61), SiT (from 22.44 to 3.83), SD3.5 (from 186.03 to 20.14), and FLUX.1 (from 196.95 to 25.67). In contrast, all models, including both one-step and multistep generation models, improved in IS, CLIP, and PickScore from one-step to 25-step sampling.

\subsection{Impact of Guidance Scale}
As shown in Table~\ref{tab:image_benchmark_performance}, one-step generation models typically use much lower default CFG values than the widely used CFG of 7.0 in SD3.5, such as MeanFlow with CFG 1 and SoFlow with CFG 2. Unsurprisingly, these low default CFG settings lead to better FID but worse IS, CLIP, and PickScore, reflecting the natural tradeoff between distributional fidelity and semantic/text alignment. Even some multistep generation models from academic research groups, including RAE, Scale RAE, and SiT, also adopt much lower default CFG values than industrial multistep models such as SD3.5 and FLUX.1. This evaluation setup may reflect the fact that FID has historically dominated image generation research, whereas in commercial image generation, text alignment and user-perceived prompt faithfulness are often more important.

\begin{table}[!t]
\centering
\scriptsize
\setlength{\tabcolsep}{1.8pt}
\renewcommand{\arraystretch}{1.12}
\makebox[\linewidth][c]{%
\begin{tabular}{l|cccccccc|cccccccc}
\toprule
& \multicolumn{8}{c|}{\textbf{ImageNetV2}} & \multicolumn{8}{c}{\textbf{reLAIONet}} \\
\cmidrule(lr){2-9}\cmidrule(lr){10-17}
\textbf{Model [steps, cfg]} &
\textbf{FID $\downarrow$} & \textbf{IS $\uparrow$} & \textbf{CLIP $\uparrow$} & \textbf{Pick $\uparrow$} &
\textbf{csFID$_2$ $\downarrow$} & \textbf{psFID$_2$ $\downarrow$} &
\textbf{csIS$_2$ $\uparrow$} & \textbf{psIS$_2$ $\uparrow$} &
\textbf{FID $\downarrow$} & \textbf{IS $\uparrow$} & \textbf{CLIP $\uparrow$} & \textbf{Pick $\uparrow$} &
\textbf{csFID$_2$ $\downarrow$} & \textbf{psFID$_2$ $\downarrow$} &
\textbf{csIS$_2$ $\uparrow$} & \textbf{psIS$_2$ $\uparrow$} \\
\midrule

Meanflow (B/4) [1, 7] & 105.70 & 13.31 & 26.62 & 18.40 & 1491.62 & 3122.05 & 0.94 & 0.45 & 118.35 & 13.37 & 26.65 & 18.40 & 1666.38 & 3495.69 & 0.95 & 0.45 \\
\textbf{Meanflow (B/4) [25, 7]} & 39.47 & \textbf{261.47} & \textbf{31.52} & \textbf{20.53} & 397.28 & 936.46 & \textbf{25.98} & \textbf{11.02} & 25.50 & \textbf{245.25} & \textbf{31.71} & \textbf{20.62} & 253.60 & 599.74 & \textbf{24.66} & \textbf{10.43} \\
Meanflow (B/4)$^{\star}$ [1, 1] & \textbf{15.65} & 122.40 & 29.36 & 19.48 & \textbf{181.55} & \textbf{412.42} & 10.55 & 4.64 & 17.85 & 121.92 & 29.56 & 19.53 & 204.28 & 467.99 & 10.65 & 4.65 \\
Meanflow (B/4)$^{\star}$ [25, 1] & 16.28 & 164.63 & 29.91 & 19.78 & 181.98 & 416.10 & 14.73 & 6.44 & \textbf{13.52} & 158.29 & 30.11 & 19.84 & \textbf{149.13} & \textbf{343.47} & 14.35 & 6.23 \\
\midrule

iMF (XL/2) [1, 7] & 14.36 & 224.85 & 30.20 & 20.15 & 157.45 & 353.68 & 20.51 & 9.13 & 10.82 & 215.89 & 30.31 & 20.21 & \textbf{117.78} & \textbf{264.91} & 19.83 & 8.82 \\
\textbf{iMF (XL/2) [25, 7]} & 17.94 & \textbf{258.59} & \textbf{30.50} & \textbf{20.44} & 192.85 & 429.40 & \textbf{24.06} & \textbf{10.80} & 11.95 & \textbf{247.92} & \textbf{30.63} & \textbf{20.52} & 127.37 & 283.80 & \textbf{23.26} & \textbf{10.44} \\
iMF (XL/2)$^{\star}$ [1, 6] & \textbf{13.91} & 219.30 & 30.14 & 20.13 & \textbf{153.12} & \textbf{343.27} & 19.92 & 8.89 & 10.64 & 211.18 & 30.27 & 20.18 & 116.12 & 261.28 & 19.35 & 8.60 \\
iMF (XL/2)$^{\star}$ [25, 6] & 17.01 & 253.26 & 30.45 & 20.40 & 183.46 & 408.74 & 23.48 & 10.54 & 11.45 & 242.27 & 30.57 & 20.48 & 122.52 & 272.99 & 22.64 & 10.16 \\
\midrule

SoFlow (XL/2)$^{\star}$ [1, 2] & \textbf{12.80} & 182.80 & 29.96 & 19.95 & \textbf{142.60} & \textbf{321.61} & 16.41 & 7.28 & \textbf{11.13} & 182.70 & 30.15 & 20.00 & \textbf{122.44} & \textbf{278.25} & 16.61 & 7.31 \\
\textbf{SoFlow (XL/2)$^{\star}$ [25, 2]} & 25.76 & \textbf{272.48} & \textbf{30.94} & \textbf{20.70} & 269.10 & 601.18 & \textbf{26.08} & \textbf{11.68} & 16.87 & \textbf{251.62} & \textbf{31.08} & \textbf{20.77} & 174.64 & 391.06 & \textbf{24.31} & \textbf{10.86} \\
\midrule

RAE [1, 7] & 176.12 & 6.60 & 24.98 & 18.88 & 2822.43 & 4940.88 & 0.41 & 0.23 & 170.29 & 6.75 & 24.85 & 18.88 & 2757.63 & 4777.33 & 0.42 & 0.24 \\
\textbf{RAE [25, 7]} & 24.29 & \textbf{282.78} & \textbf{31.30} & \textbf{20.59} & 247.94 & 572.95 & \textbf{27.70} & \textbf{11.99} & 15.02 & \textbf{259.25} & \textbf{31.51} & \textbf{20.67} & 151.28 & 351.55 & \textbf{25.74} & \textbf{11.08} \\
RAE$^{\star}$ [1, 1.42] & 305.88 & 3.44 & 24.08 & 18.71 & 5275.19 & 8737.83 & 0.20 & 0.12 & 296.52 & 3.36 & 24.06 & 18.75 & 5122.27 & 8434.35 & 0.20 & 0.12 \\
RAE$^{\star}$ [25, 1.42] & \bestblue{11.86} & 199.58 & 30.31 & 20.44 & \bestblue{129.10} & \bestblue{283.87} & 18.34 & 8.34 & 10.79 & 193.51 & 30.47 & 20.49 & 116.22 & 257.00 & 17.97 & 8.12 \\
\midrule

Scale RAE [1, 7] & 316.03 & 1.53 & 20.39 & 16.89 & 7601.41 & 11078.20 & 0.06 & 0.04 & 319.00 & 1.53 & 20.38 & 16.90 & 7680.37 & 11169.08 & 0.06 & 0.04 \\
Scale RAE [25, 7] & 26.99 & 78.80 & \textbf{30.03} & \textbf{21.06} & 299.29 & 608.54 & 7.11 & 3.50 & \textbf{22.91} & \textbf{86.23} & \textbf{30.34} & \textbf{21.19} & \textbf{248.88} & \textbf{510.23} & \textbf{7.94} & \textbf{3.87} \\
Scale RAE$^{\star}$ [1, 1.42] & 316.37 & 1.53 & 20.40 & 16.89 & 7602.12 & 11090.11 & 0.06 & 0.04 & 318.57 & 1.53 & 20.38 & 16.90 & 7670.02 & 11154.02 & 0.06 & 0.04 \\
\textbf{Scale RAE$^{\star}$ [25, 1.42]} & \textbf{26.69} & \textbf{80.37} & 30.01 & \textbf{21.06} & \textbf{296.36} & \textbf{601.77} & \textbf{7.24} & \textbf{3.56} & 22.92 & 85.29 & \textbf{30.34} & \textbf{21.19} & 248.99 & 510.45 & 7.85 & 3.83 \\
\midrule

SiT ODE [Dyn, 7] & \textbf{12.99} & 196.58 & 29.99 & 20.16 & \textbf{144.43} & \textbf{319.62} & 17.68 & 7.99 & 23.07 & \bestblue{309.06} & \textbf{31.56} & \textbf{20.94} & 231.62 & 526.13 & \bestblue{30.78} & \bestblue{13.55} \\
\textbf{SiT ODE$^{\star}$ [Dyn, 1.5]} & 38.08 & \bestblue{331.79} & \textbf{31.34} & \textbf{20.84} & 387.70 & 876.80 & \bestblue{32.59} & \bestblue{14.41} & \bestblue{10.18} & 190.23 & 30.18 & 20.24 & \bestblue{111.77} & \bestblue{248.50} & 17.33 & 7.79 \\
\midrule

SD3.5 Large [1, 7] & 182.59 & 2.71 & 24.43 & 18.31 & 3059.36 & 5446.28 & 0.16 & 0.09 & 178.19 & 2.68 & 24.39 & 18.31 & 2995.43 & 5315.04 & 0.16 & 0.09 \\
SD3.5 Large [25, 7] & \textbf{31.55} & 161.51 & 31.95 & 21.78 & 309.07 & 665.10 & 16.49 & 7.66 & 21.12 & 171.48 & 32.12 & 21.85 & 204.71 & 442.38 & 17.69 & 8.19 \\
SD3.5 Large$^{\star}$ [1, 3.5] & 190.13 & 2.66 & 24.52 & 18.60 & 3162.35 & 5495.72 & 0.16 & 0.09 & 181.90 & 2.64 & 24.52 & 18.61 & 3025.46 & 5252.18 & 0.16 & 0.09 \\
\textbf{SD3.5 Large$^{\star}$ [25, 3.5]} & 31.69 & \textbf{171.34} & \bestblue{32.10} & \textbf{22.05} & \textbf{307.55} & \textbf{651.79} & \textbf{17.66} & \textbf{8.33} & \textbf{20.60} & \textbf{180.44} & \bestblue{32.26} & \textbf{22.13} & \textbf{197.94} & \textbf{420.63} & \textbf{18.78} & \textbf{8.84} \\
\midrule

Flux.1 [1, 7] & 184.82 & 3.95 & 24.51 & 18.90 & 3076.54 & 5173.99 & 0.24 & 0.14 & 182.46 & 4.09 & 24.69 & 18.95 & 2993.13 & 5081.00 & 0.25 & 0.15 \\
Flux.1 [25, 7] & 36.24 & 99.05 & 30.39 & 21.79 & 392.40 & 763.26 & 9.15 & 4.70 & 29.29 & 108.79 & 30.73 & 21.90 & 310.17 & 610.71 & 10.27 & 5.22 \\
Flux.1$^{\star}$ [1, 3.5] & 198.74 & 3.86 & 25.04 & 19.11 & 3169.69 & 5442.07 & 0.24 & 0.14 & 200.32 & 3.99 & 25.17 & 19.15 & 3161.97 & 5462.44 & 0.25 & 0.15 \\
\textbf{Flux.1$^{\star}$ [25, 3.5]} & \textbf{34.63} & \textbf{125.78} & \textbf{30.96} & \bestblue{22.13} & \textbf{361.29} & \textbf{707.11} & \textbf{12.06} & \textbf{6.16} & \textbf{26.45} & \textbf{131.72} & \textbf{31.24} & \bestblue{22.24} & \textbf{271.02} & \textbf{534.76} & \textbf{12.86} & \textbf{6.52} \\

\bottomrule
\end{tabular}}
\caption{ImageNetV2 and reLAIONet benchmarks. $^{\star}$ denotes model-specific tuned CFG. SoFlow prevents custom CFG and SiT prevents explicit step control. We report FID, IS, CLIP, Pick, and semantics-scaled metrics csFID$_2$, psFID$_2$, csIS$_2$, and psIS$_2$. Bold marks the best within each model family. Blue marks the best value across all models for each metric.}
\label{tab:imagenetv2_relaionet_benchmark_performance}
\end{table}

\subsection{Text-Conditioned vs.\ Class-Conditioned Models}
From Table~\ref{tab:image_benchmark_performance}, it can be observed that class-conditioned models typically have much lower FID values (can be lower than 10), whereas text-conditioned models have FIDs around 20. Correspondingly, text-conditioned models generally have better IS, CLIP, and PickScore. This split arises because label-conditioned models are trained directly on ImageNet classes, so their output distribution is optimized to match the ImageNet reference set, while text-conditioned models trained on LAION-style data produce outputs better aligned with human preference. Ranking models on FID alone would therefore declare label-conditioned models uniformly superior, hiding that SD3.5~Large achieves the highest CLIP (32.10) and FLUX.1 the highest PickScore (22.13) across all models, hence motivating reporting all metrics under matched settings.

\subsection{Generalization Under Distribution Shift: ImageNetV2 and reLAIONet}
Table~\ref{tab:imagenetv2_relaionet_benchmark_performance} evaluates models under distribution shift using reLAIONet. Class-conditioned models show a substantial drop in their FID: RAE goes from FID 2.61 on ImageNet to 10.79 on reLAIONet, indicating that its strong ImageNet FID shows overfitting to the reference distribution rather than general generation quality. Text-conditioned models remain stable in CLIP and PickScore across all three benchmarks: SD3.5 Large holds CLIP at 32.10, 32.10, and 32.26 across ImageNet, ImageNetV2, and reLAIONet, and FLUX.1 maintains PickScore at 22.13, 22.13 and 22.24, consistent with the broader diversity of their pretraining data. csFID and psFID amplify this signal further, with the gap between raw and semantics-scaled FID widening most for class-conditioned models on reLAIONet, directly exposing overfitting that raw FID conceals.

\subsection{Semantics-Scaled Metrics}
The semantics-scaled metrics, including csFID, psFID, csIS, and psIS, incorporate semantic alignment into FID and IS, thereby making the resulting scores more comparable across different sampling and guidance settings. For example, on ImageNetV2, SD3.5 Large [25, 7] achieves a slightly better FID than SD3.5 Large [25, 3.5] (31.55 vs. 31.69), but has a worse CLIP score (31.95 vs. 32.10). After applying the semantics-scaled metrics, however, SD3.5 Large [25, 3.5] is considered superior according to both csFID and psFID.

\subsection{Guidance Scale and Step Ablation on MeanFlow}
\begin{figure}[!t]
    \centering
    \makebox[\linewidth][c]{%
    \includegraphics[width=1\textwidth]{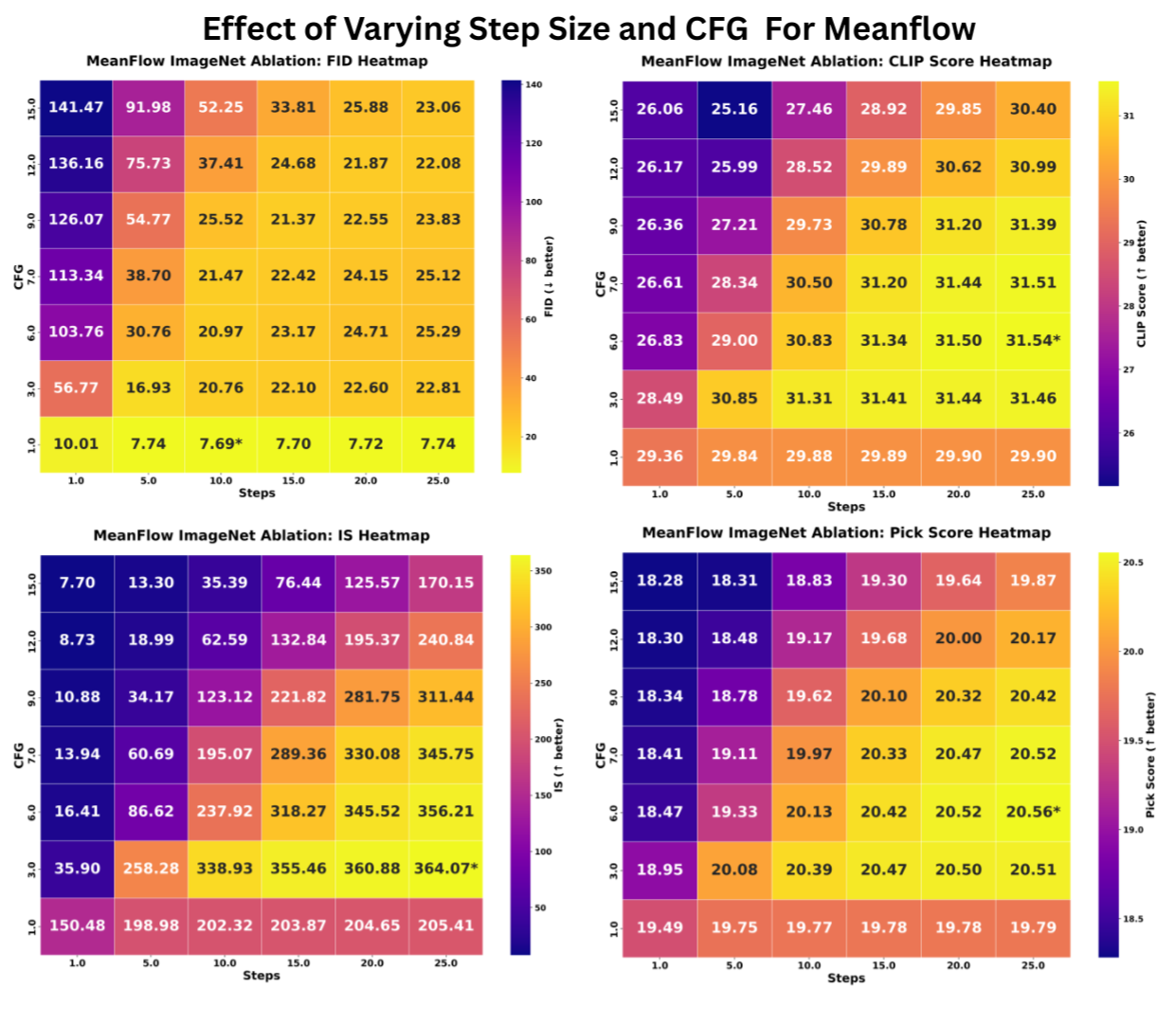}
    }
    \caption{Effect of step size and CFG on MeanFlow performance. Heatmaps show FID, CLIP Score, Inception Score, and PickScore across different step sizes and CFG values}
    \label{fig:ablation_heatmap}
\end{figure}

Figure~\ref{fig:ablation_heatmap} shows heatmaps of FID, IS, CLIP, and PickScore for MeanFlow across a full grid of CFG values and step counts. FID decreases monotonically as CFG approaches 1.0 at every step count, while IS and PickScore increase with both CFG and steps. The CFG that minimizes FID is precisely the one that lowers IS, CLIP, and PickScore. Since PickScore is trained as a proxy for human preference, this means FID-only model selection favors outputs that human observers rate lower. Hence, jointly reporting IS, CLIP, and PickScore alongside FID is needed to make this tradeoff visible.

\subsection{Qualitative Comparison Across Models}
Figures~\ref{fig:within_qualitative_results} and~\ref{fig:qualitative_results} show that Scale~RAE, FLUX.1, and SD3.5~Large produce more coherent outputs with fewer visible distortions, consistent with their leading PickScore ($\geq$21.7) and CLIP (
$\geq$30.9). MeanFlow and iMF exhibit visible distortions in faces, scuba-diver structure, and fine detail despite FID values as low as 3.85 and 7.74, confirming that low FID does not imply high visual quality. csFID, psFID, csIS, and psIS reflect these failures more clearly than raw FID, as they weight FID and IS by CLIP and PickScore. The qualitative ranking, with Scale~RAE, FLUX.1, and SD3.5~Large ahead of MeanFlow, iMF, and RAE, aligns with PickScore and CLIP rather than FID, reinforcing that the full metric suite is necessary to surface the true quality ordering.
\section{Discussion}
\subsection{Quantitative and Qualitative Mismatch}
One-step models become quantitatively competitive with multi-step systems at 25-step inference, yet this numerical parity does not translate to visual quality. As seen in Figures~\ref{fig:within_qualitative_results} and~\ref{fig:qualitative_results}, one-step models still exhibit characteristic local distortions in faces and fine object structures that multi-step systems such as SD3.5~Large and FLUX.1 avoid entirely. This mismatch is precisely what FID-only evaluation obscures: a model can score competitively on FID while producing samples that are clearly inferior to human observers, as confirmed by lower PickScore and CLIP Score. Closing the gap between one-step and multi-step generators therefore requires jointly optimizing FID, IS, CLIP Score, and PickScore rather than FID alone.
 
\subsection{Limitations}
The evaluated models differ in training data, conditioning interface, and intended inference setting, which limits direct comparison. Label-ID-conditioned models are optimized against the ImageNet class interface, giving them a structural FID advantage over text-conditioned models such as SD3.5~Large and FLUX.1, whose weaker FID on ImageNet benchmarks reflects a conditioning mismatch rather than inferior general generation ability. reLAIONet partially addresses this by shifting the reference distribution, but a fully fair comparison would require either a shared conditioning interface or benchmarks designed specifically for each model's training paradigm. We also note that PickScore, while a strong proxy for human preference, was trained on a fixed set of human judgments and may not generalize uniformly across all image domains and styles represented in our benchmark.

\section{Conclusion}
We present a setting-matched and semantics-scaled benchmark comparing one-step and multi-step generative models across ImageNet, ImageNetV2, and reLAIONet, organized around three findings. First, fair comparison of one-step and multi-step generators requires matched step and guidance settings: step count improves IS, CLIP Score, and PickScore, while CFG primarily governs FID, and mixing these axes produces misleading rankings. Second, label-ID-conditioned models overfit to the ImageNet reference distribution, and out-of-distribution benchmarks such as reLAIONet, which preserve the class interface while replacing the validation images, are necessary to distinguish genuine generation quality from distributional memorization. Text-conditioned models such as SD3.5~Large and Flux.1-dev remain stable across all three datasets, whereas class-conditioned models such as RAE show a substantial FID degradation under the shift. Third, FID-only evaluation hides the tradeoff between distributional matching and perceptual quality: models such as MeanFlow and iMF achieve low FID while producing visible distortions in faces and fine structures, a failure that our semantics-scaled metrics, csFID, psFID, csIS, and psIS, make explicit by separating class-level distributional quality from semantic and perceptual coherence. Together, these three points motivate evaluating label-ID-conditioned generators with the full metric suite alongside reLAIONet, and future work on one-step models should optimize toward csFID, psFID, csIS, and psIS rather than FID alone.

\bibliographystyle{plainnat}
\bibliography{main}

\newpage

\appendix
\section{Additional Across Family Qualitative Samples}

\begin{figure}[H]
    \centering
    \makebox[\linewidth][c]{%
    \includegraphics[width=1\textwidth]{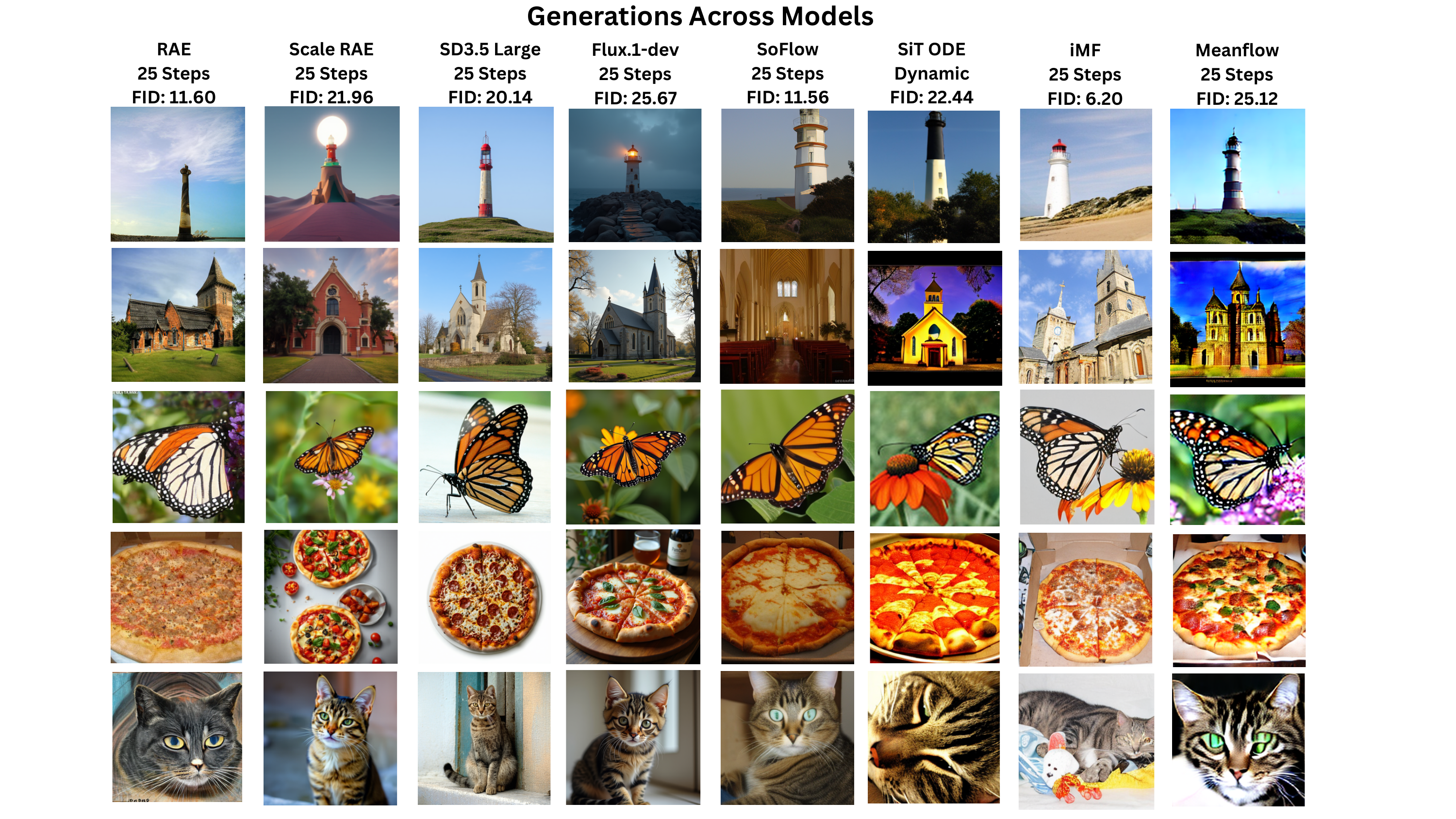}
    }
    \caption{Comparison of generation quality across all models. Images generated with classes lighthouse, church, butterfly, pizza and cat. Models settings are 25 steps and CFG 7}
    \label{fig:qualitative_results_extra}
\end{figure}

Figure~\ref{fig:qualitative_results_extra} compares model generations for additional classes under the matched setting of 25 steps and CFG 7. The results reinforce the same trend as the main qualitative figure: stronger FID does not always imply stronger visual quality. One-step models such as MeanFlow, iMF, and SoFlow can achieve competitive FID values, but still show local artifacts in fine structures, object boundaries, and texture consistency, especially in the lighthouse, butterfly, and cat examples. In contrast, SD3.5 and Flux often produce more coherent samples despite weaker FID, further showing that FID-only evaluation can miss perceptual quality differences captured by csFID, psFID, csIS, and psIS.

\section{Reported CFG and Step Size for Each Approach}
\begin{table}[H]
\centering
\small
\label{tab:model_settings}
\begin{tabular}{lcccp{4.2cm}}
\toprule
\textbf{Model} & \textbf{Reported Inference CFG} & \textbf{Reported Steps} & \textbf{Model Weights} \\
\midrule
MeanFlow    & 1.0  & 1      & DiT-B/4 \\
iMF         & 6.0  & 1      & DiT-XL/2 \\
SoFlow      & 2.0  & 1      & DiT-X/2-cond \\
RAE         & 1.42 & 50      & DiT-XL, DINOv2-B \\
Scale-RAE   & 1.42 & 50      & Qwen1.5B\_DiT2.4B \\
SiT (ODE)   & 1.5  & Dynamic    & SiT-XL/2 \\
SD3.5-Large & 3.5  & 50      & stable-diffusion-3.5-large \\
FLUX.1-dev  & 3.5  & 50      & FLUX.1-dev \\
\bottomrule
\end{tabular}
\caption{Models evaluated, the checkpoints used, and their reported guidance/step settings from original releases. The original release CFG values for SD3.5-Large and FLUX.1-dev were taken based on model cards from HuggingFace, and the original step count for FLUX.1-dev was also taken from the model card}
\end{table}

\section{Re-implementation Performance vs. Reported Performance For all Models}

\begin{table}[H]
\centering
\setlength{\tabcolsep}{4pt} 
\renewcommand{\arraystretch}{1.1} 
\makebox[\linewidth][c]{%
\begin{tabular}{llcccccc}
\toprule
\textbf{Dataset} & \textbf{Model (Natural Setting)} &
\textbf{NFE} & \textbf{CFG} &
\textbf{FID $\downarrow$} & \textbf{IS $\uparrow$} & \textbf{CLIP Score $\uparrow$} &
\textbf{Pick Score $\uparrow$} \\
\midrule

\multirow{16}{*}{Imagenet}
& Meanflow (B/4) (Ours) & 1 & 1      & 10.00   & 150.47 & 29.36 & 19.49 \\
& Meanflow (B/4) (Original) & 1 & 1      & 11.40   & - & - & - \\
\cmidrule(lr){2-8}

& iMF (XL/2) (Ours) & 1 & 6         & 3.85   & 285.50 & 30.15 & 20.13 \\
& iMF (XL/2) (Original) & 1 & 6     & 1.70   & 282.00 & - &  - \\

\cmidrule(lr){2-8}
& SoFlow (Ours) & 1 & 2        & 3.97   & 238.97 & 29.98 & 19.95 \\
& SoFlow (Original) & 1 & 2        &  2.96  & - & - & - \\

\cmidrule(lr){2-8}
& RAE (Ours) & 50 & 1.42         & 2.86   & 265.13 & 30.32 & 20.46 \\
& RAE (Original) & 50 & 1.42         &  2.16 & - & - & - \\

\cmidrule(lr){2-8}
& SiT ODE (Ours) & 125 & 1.5  & 3.83   & 254.85 & 30.00 & 20.17 \\
& SiT ODE (Original) & 125 & 1.5  & 2.15   & 258.09 & - & - \\

\midrule
\bottomrule
\end{tabular}
}
\caption{Re-implementation of each model and how it matches up to the original work. Difference arise due to different random seed and using JAX + GPU whereas implementations use JAX + TPU. Additional variations emerge as methods might use ImageNet train set instead of validation set. Blanks refer to no official results.  Scale RAE, SD3.5 and FLUX.1 have no officially reported FID or other metrics we can compare with. For SD3.5-large, similar setting downstream comparison by Stability AI reports FID of 20.06, and CLIP Score 32.00 and ours reports FID 20.14 and CLIP Score 32.10}
\label{tab:image_benchmark_performance}
\end{table}

\section{SiT ODE Dynamic Step Size Distribution}
\begin{table}[H]
\centering

\label{tab:sit_ode_nfe}
\makebox[\linewidth][c]{%
\begin{tabular}{lrrrrrrrr}
\toprule
Dataset &  CFG & NFE Mean & NFE Min & NFE Max & FID$\downarrow$ & IS$\uparrow$ & CLIP$\uparrow$ & Pick$\uparrow$ \\
\midrule
\multirow{2}{*}{ImageNet}
 &  7.00 & 155.95 &  92 & 284 & 22.44 & 466.54 & 31.34 & 20.84 \\
 & 1.50 &  63.13 &  44 & 116 &  3.83 & 254.85 & 30.00 & 20.17 \\
\cmidrule(lr){2-9}
\multirow{2}{*}{ImageNet-v2}
 &  7.00 & 158.57 & 104 & 266 & 38.08 & 331.80 & 31.34 & 20.84 \\
 &  1.50 &  63.43 &  44 &  98 & 12.99 & 196.58 & 30.00 & 20.16 \\
\cmidrule(lr){2-9}
\multirow{2}{*}{RelAIONet}
 &  7.00 & 156.02 &  92 & 302 & 23.07 & 309.06 & 31.55 & 20.93 \\
 &  1.50 &  63.25 &  44 &  98 & 10.18 & 190.23 & 30.18 & 20.24 \\
\bottomrule
\end{tabular}
}
\caption{SiT ODE results with adaptive NFE across datasets. Shows the step sizes dynamic spread. For SiT, 2 NFE = 1 step size}
\end{table}

\section{Additional Within Family Qualitative Samples}
All examples show images generated with classes Golden Retriever, GoldFish, Groom, Scuba Diver and Sports Car with images ordered in ascending order of MinMax Harmonic Mean (MMHM).
\subsection{Within Family for FLUX.1-dev and SD3.5}
\begin{figure}[H]
    \centering
    \makebox[\linewidth][c]{%
    \includegraphics[width=0.9\textwidth]{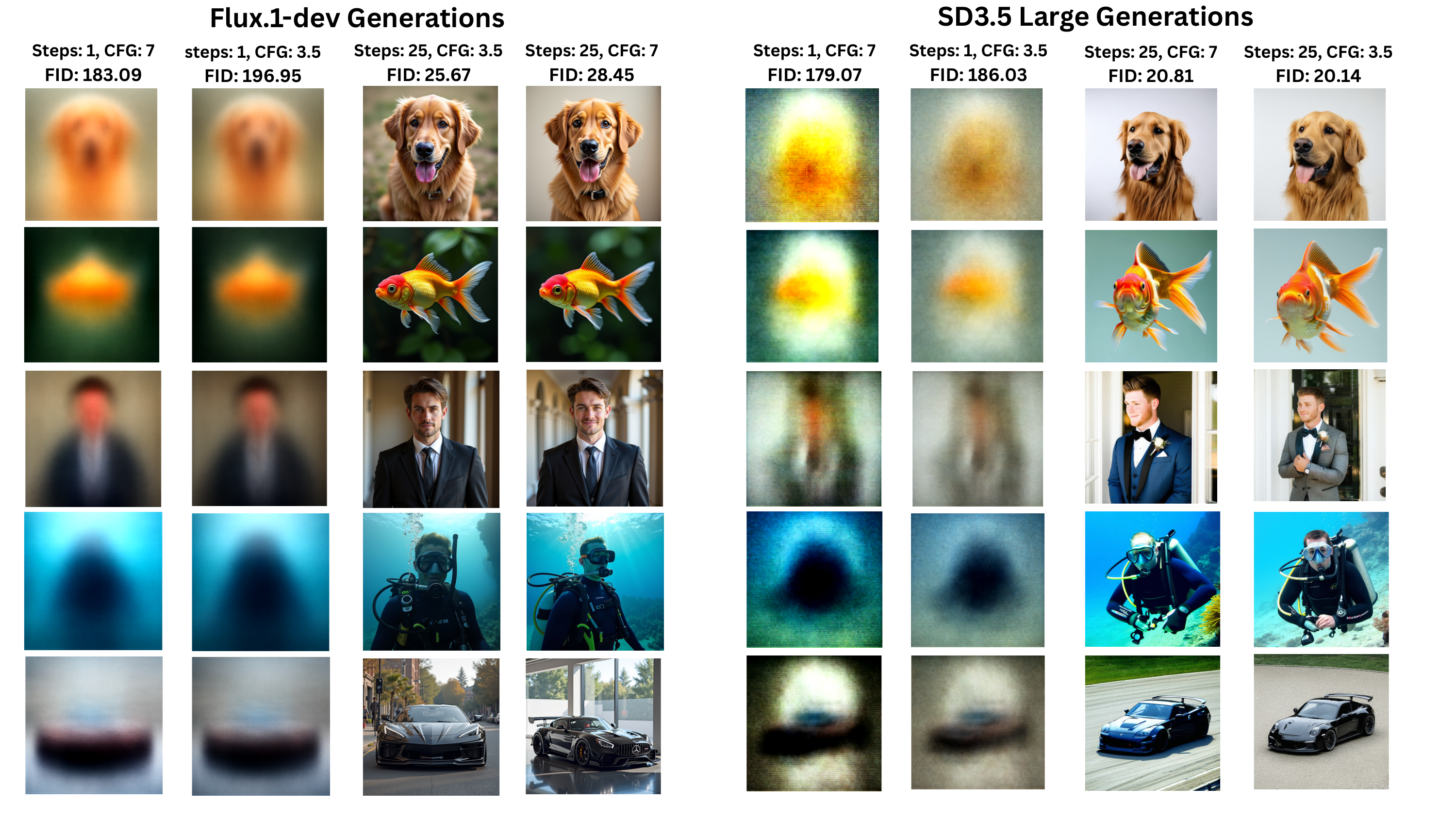}
    }
    \caption{Within model family comparison for FLUX.1-dev and Stable Diffusion 3.5 Large}
\end{figure}

\subsection{Within Family for Scale RAE and RAE}
\begin{figure}[H]
    \centering
    \makebox[\linewidth][c]{%
    \includegraphics[width=0.9\textwidth]{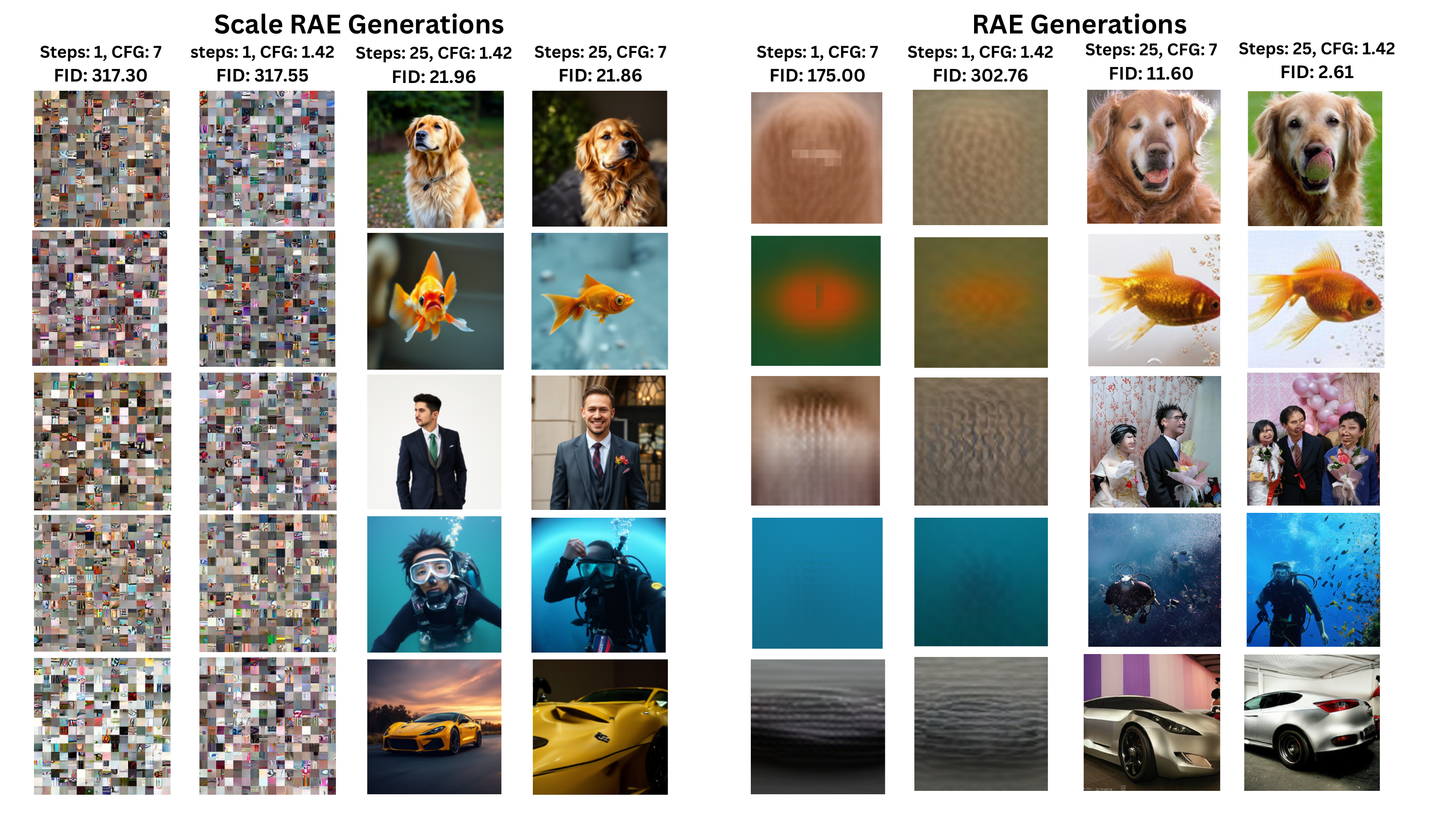}
    }
    \caption{Within model family comparison for Scale RAE and RAE}
\end{figure}

\subsection{Within Family for SiT ODE and SoFlow}
\begin{figure}[H]
    \centering
    \makebox[\linewidth][c]{%
    \includegraphics[width=1\textwidth]{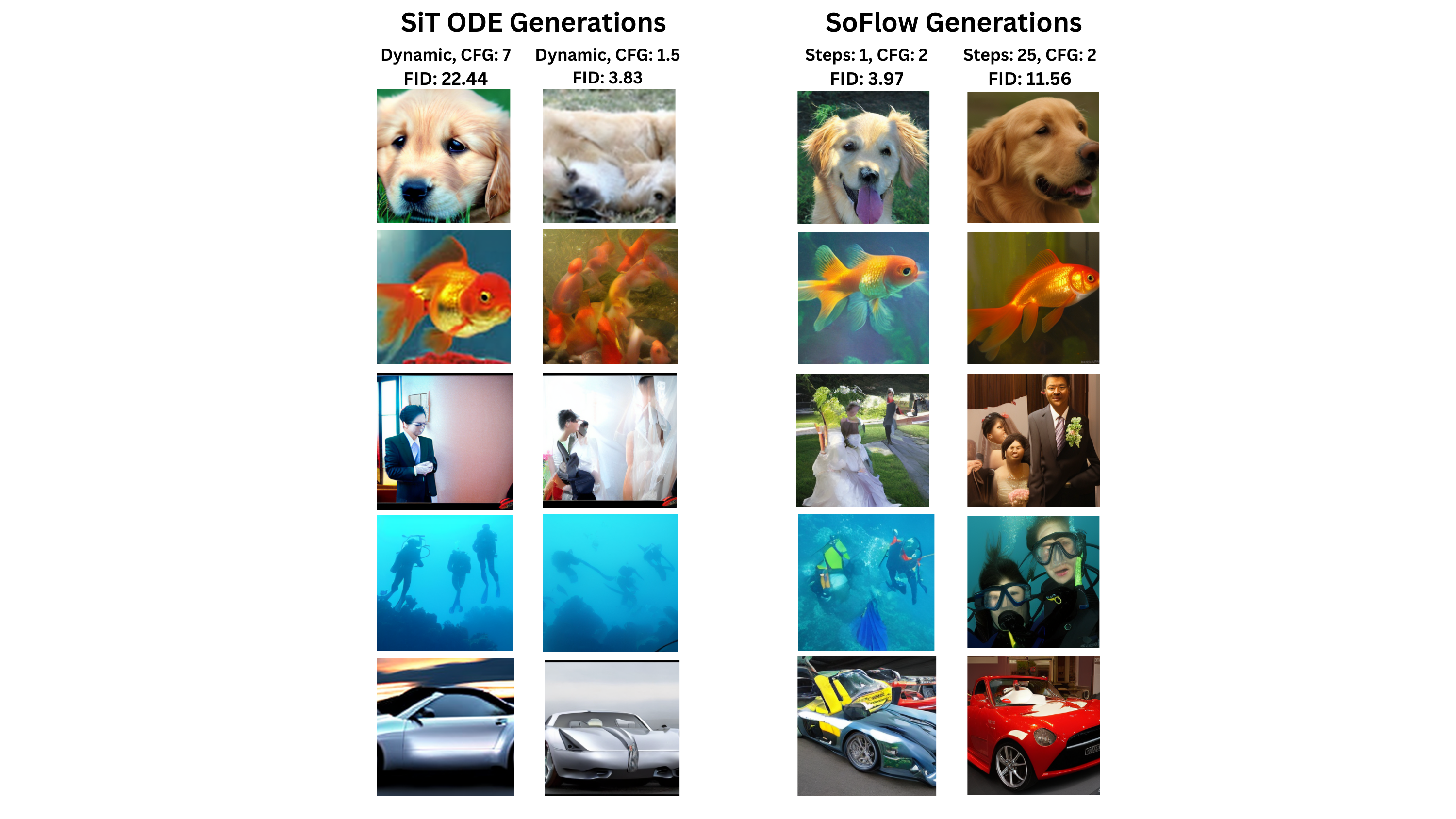}
    }
    \caption{Within model family comparison for SoFlow and SiT ODE}
\end{figure}

\section{Metric Correlation Matrix}
\begin{figure}[H]
    \centering
    \makebox[\linewidth][c]{%
    \includegraphics[width=1\textwidth]{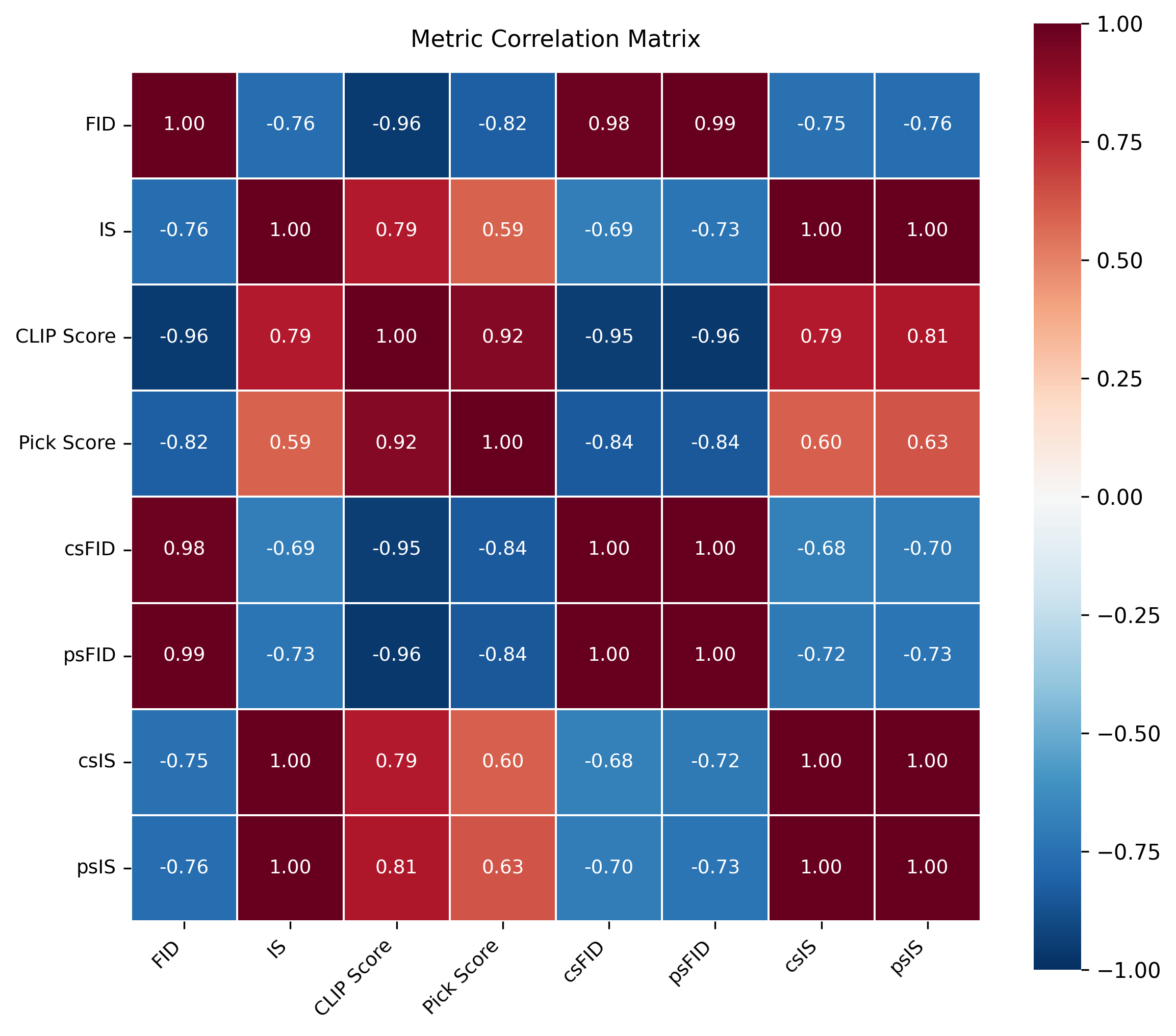}
    }
    \caption{Correlation matrix between base and scaled metrics. csFID and psFID remain strongly correlated with FID while incorporating negative dependence on CLIP and PickScore, whereas csIS and psIS track IS while preserving semantic and preference alignment. This shows that the scaled metrics retain the behavior of the base metrics while making alignment and preference tradeoffs explicit.}
\end{figure}
The correlation matrix shows that the scaled metrics preserve the intended structure of the base metrics while making the FID-alignment tradeoff explicit. csFID and psFID are highly correlated with FID, meaning they still capture distributional quality, but they are strongly negatively correlated with CLIP Score and PickScore, showing that they penalize low-alignment or low-preference generations. Similarly, csIS and psIS are nearly perfectly correlated with IS, but also retain positive correlation with CLIP and PickScore. This supports using csFID, psFID, csIS, and psIS as diagnostics rather than replacements: they expose when improvements in FID come at the cost of semantic alignment or perceptual quality.

\section{Alpha Variation for csFID, psFID, csIS and psIS}
\subsection{ImageNet Validation}

\begin{table}[!h]
\centering
\scriptsize
\setlength{\tabcolsep}{1.2pt}
\renewcommand{\arraystretch}{1.08}
\makebox[\linewidth][c]{%
\begin{tabular}{lcccccccccccccc}
\toprule
\textbf{Model} &
\textbf{Steps} & \textbf{CFG} &
\textbf{FID $\downarrow$} & \textbf{IS $\uparrow$} &
\textbf{CLIP $\uparrow$} & \textbf{Pick $\uparrow$} &
\textbf{csFID$_1$ $\downarrow$} & \textbf{psFID$_1$ $\downarrow$} &
\textbf{csIS$_1$ $\uparrow$} & \textbf{psIS$_1$ $\uparrow$} &
\textbf{csFID$_2$ $\downarrow$} & \textbf{psFID$_2$ $\downarrow$} &
\textbf{csIS$_2$ $\uparrow$} & \textbf{psIS$_2$ $\uparrow$} \\
\midrule

Meanflow (B/4) & 1 & 7.0 & 113.34 & 13.93 & 26.61 & 18.41 & 425.93 & 615.64 & 3.71 & 2.56 & 1600.64 & 3344.07 & 0.99 & 0.47 \\
\textbf{Meanflow (B/4)} & 25 & 7.0 & 25.12 & \textbf{346.19} & \textbf{31.51} & \textbf{20.52} & 79.72 & 122.42 & \textbf{109.08} & \textbf{71.04} & 253.00 & 596.58 & \textbf{34.37} & \textbf{14.58} \\
Meanflow (B/4)$^{\star}$ & 1 & 1.0 & 10.00 & 150.47 & 29.36 & 19.49 & 34.06 & 51.31 & 44.18 & 29.33 & 116.01 & 263.25 & 12.97 & 5.72 \\
Meanflow (B/4)$^{\star}$ & 25 & 1.0 & \textbf{7.74} & 205.27 & 29.90 & 19.76 & \textbf{25.89} & \textbf{39.17} & 61.38 & 40.56 & \textbf{86.58} & \textbf{198.23} & 18.35 & 8.02 \\
\midrule

iMF (XL/2) & 1 & 7.0 & 4.07 & 293.18 & 30.20 & 20.15 & 13.48 & 20.20 & 88.54 & 59.08 & 44.62 & 100.24 & 26.74 & 11.90 \\
\textbf{iMF (XL/2)} & 25 & 7.0 & 6.20 & \textbf{346.87} & \textbf{30.49} & \textbf{20.45} & 20.33 & 30.32 & \textbf{105.76} & \textbf{70.93} & 66.69 & 148.25 & \textbf{32.25} & \textbf{14.51} \\
iMF (XL/2)$^{\star}$ & 1 & 6.0 & \textbf{3.85} & 285.50 & 30.15 & 20.13 & \textbf{12.77} & \textbf{19.13} & 86.08 & 57.47 & \textbf{42.35} & \textbf{95.01} & 25.95 & 11.57 \\
iMF (XL/2)$^{\star}$ & 25 & 6.0 & 5.62 & 336.84 & 30.43 & 20.41 & 18.47 & 27.54 & 102.50 & 68.75 & 60.69 & 134.91 & 31.19 & 14.03 \\
\midrule

SoFlow (XL/2)$^{\star}$ & 1 & 2.0 & \textbf{3.97} & 238.98 & 29.98 & 19.95 & \textbf{13.24} & \textbf{19.90} & 71.65 & 47.68 & \textbf{44.17} & \textbf{99.75} & 21.48 & 9.51 \\
\textbf{SoFlow (XL/2)$^{\star}$} & 25 & 2.0 & 11.56 & \textbf{362.88} & \textbf{30.95} & \textbf{20.72} & 37.35 & 55.79 & \textbf{112.31} & \textbf{75.19} & 120.68 & 269.26 & \textbf{34.76} & \textbf{15.58} \\
\midrule

RAE & 1 & 7.0 & 175.00 & 6.69 & 24.98 & 18.88 & 700.56 & 926.91 & 1.67 & 1.26 & 2804.49 & 4909.46 & 0.42 & 0.24 \\
\textbf{RAE} & 25 & 7.0 & 11.60 & \textbf{382.36} & \textbf{31.33} & \textbf{20.58} & 37.03 & 56.37 & \textbf{119.79} & \textbf{78.69} & 118.18 & 273.88 & \textbf{37.53} & \textbf{16.19} \\
RAE$^{\star}$ & 1 & 1.42 & 302.76 & 3.43 & 24.08 & 18.71 & 1257.31 & 1618.17 & 0.83 & 0.64 & 5221.38 & 8648.70 & 0.20 & 0.12 \\
RAE$^{\star}$ & 25 & 1.42 & \bestblue{2.61} & 255.53 & 30.31 & 20.43 & \bestblue{8.61} & \bestblue{12.78} & 77.45 & 52.20 & \bestblue{28.41} & \bestblue{62.53} & 23.48 & 10.66 \\
\midrule

Scale RAE & 1 & 7.0 & 317.30 & 1.53 & 20.39 & 16.89 & 1556.15 & 1878.63 & 0.31 & 0.26 & 7631.95 & 11122.71 & 0.06 & 0.04 \\
Scale RAE & 25 & 7.0 & \textbf{21.86} & \textbf{90.49} & 30.01 & 21.05 & \textbf{72.84} & \textbf{103.85} & \textbf{27.16} & \textbf{19.05} & \textbf{242.73} & \textbf{493.34} & 8.15 & \textbf{4.01} \\
Scale RAE$^{\star}$ & 1 & 1.42 & 317.55 & 1.53 & 20.39 & 16.89 & 1557.38 & 1880.11 & 0.31 & 0.26 & 7637.97 & 11131.48 & 0.06 & 0.04 \\
\textbf{Scale RAE$^{\star}$} & 25 & 1.42 & 21.96 & 90.40 & \textbf{30.03} & \textbf{21.06} & 73.13 & 104.27 & 27.15 & 19.04 & 243.51 & 495.13 & \textbf{8.15} & 4.01 \\
\midrule

\textbf{SiT ODE} & Dynamic & 7.0 & 22.44 & \bestblue{466.54} & \textbf{31.34} & \textbf{20.84} & 71.60 & 107.68 & \bestblue{146.21} & \bestblue{97.23} & 228.47 & 516.69 & \bestblue{45.82} & \bestblue{20.26} \\
SiT ODE$^{\star}$ & Dynamic & 1.5 & \textbf{3.83} & 254.45 & 30.00 & 20.17 & \textbf{12.77} & \textbf{18.99} & 76.33 & 51.32 & \textbf{42.56} & \textbf{94.14} & 22.90 & 10.35 \\
\midrule

SD3.5 Large & 1 & 7.0 & 179.07 & 2.71 & 24.42 & 18.31 & 733.29 & 977.99 & 0.66 & 0.50 & 3002.84 & 5341.29 & 0.16 & 0.09 \\
SD3.5 Large & 25 & 7.0 & 20.81 & 202.15 & 31.96 & 21.78 & 65.11 & 95.55 & 64.61 & 44.03 & 203.73 & 438.69 & 20.65 & 9.59 \\
SD3.5 Large$^{\star}$ & 1 & 3.5 & 186.03 & 2.68 & 24.52 & 18.60 & 758.69 & 1000.16 & 0.66 & 0.50 & 3094.16 & 5377.21 & 0.16 & 0.09 \\
\textbf{SD3.5 Large$^{\star}$} & 25 & 3.5 & \textbf{20.14} & \textbf{214.58} & \bestblue{32.10} & \textbf{22.04} & \textbf{62.74} & \textbf{91.38} & \textbf{68.88} & \textbf{47.29} & \textbf{195.46} & \textbf{414.61} & \textbf{22.11} & \textbf{10.42} \\
\midrule

Flux.1-dev & 1 & 7.0 & 183.09 & 3.97 & 24.49 & 18.90 & 747.61 & 968.73 & 0.97 & 0.75 & 3052.72 & 5125.56 & 0.24 & 0.14 \\
Flux.1-dev & 25 & 7.0 & 28.45 & 115.62 & 30.36 & 21.79 & 93.71 & 130.56 & 35.10 & 25.19 & 308.66 & 599.19 & 10.66 & 5.49 \\
Flux.1-dev$^{\star}$ & 1 & 3.5 & 196.95 & 3.89 & 25.01 & 19.10 & 787.49 & 1031.15 & 0.97 & 0.74 & 3148.68 & 5398.70 & 0.24 & 0.14 \\
\textbf{Flux.1-dev$^{\star}$} & 25 & 3.5 & \textbf{25.67} & \textbf{148.98} & \textbf{30.96} & \bestblue{22.13} & \textbf{82.91} & \textbf{116.00} & \textbf{46.12} & \textbf{32.97} & \textbf{267.81} & \textbf{524.16} & \textbf{14.28} & \textbf{7.30} \\

\bottomrule
\end{tabular}
}
\caption{ImageNet validation benchmark comparing scaled metrics with $\alpha=1$ and $\alpha=2$. $^{\star}$ denotes model-specific tuned CFG. SoFlow prevents custom CFG and SiT prevents explicit step control. Bold marks the best within each model family. Blue marks the best value across all models for each metric.}
\label{tab:image_benchmark_alpha_comparison}
\end{table}

Table~\ref{tab:image_benchmark_alpha_comparison} compares $\alpha=1$ and $\alpha=2$ for the scaled metrics. Both choices preserve the same broad trends, but $\alpha=1$ keeps the scaled metrics close to the base FID and IS because CLIP Score and PickScore vary over a narrow range. Using $\alpha=2$ stretches these compressed alignment and preference signals, making the FID-quality tradeoff more visible while preserving stable rankings. In particular, low-FID settings still achieve strong csFID$_2$ and psFID$_2$, but csIS$_2$ and psIS$_2$ more clearly expose when semantic or perceptual quality does not improve. Thus, we use $\alpha=2$ in the main paper as a stronger diagnostic while reporting $\alpha=1$ in the supplement for sensitivity.

\pagebreak
\subsection{ImageNetv2}

\begin{table}[!h]
\centering
\scriptsize
\setlength{\tabcolsep}{0.9pt}
\renewcommand{\arraystretch}{1.08}
\makebox[\linewidth][c]{%
\begin{tabular}{lcccccccccccccc}
\toprule
\textbf{Model} &
\textbf{Steps} & \textbf{CFG} &
\textbf{FID $\downarrow$} & \textbf{IS $\uparrow$} &
\textbf{CLIP $\uparrow$} & \textbf{Pick $\uparrow$} &
\textbf{csFID$_1$ $\downarrow$} & \textbf{psFID$_1$ $\downarrow$} &
\textbf{csIS$_1$ $\uparrow$} & \textbf{psIS$_1$ $\uparrow$} &
\textbf{csFID$_2$ $\downarrow$} & \textbf{psFID$_2$ $\downarrow$} &
\textbf{csIS$_2$ $\uparrow$} & \textbf{psIS$_2$ $\uparrow$} \\
\midrule

Meanflow (B/4) & 1 & 7.0 & 105.70 & 13.31 & 26.62 & 18.40 & 397.07 & 574.46 & 3.54 & 2.45 & 1491.62 & 3122.05 & 0.94 & 0.45 \\
Meanflow (B/4) & 25 & 7.0 & 39.47 & \textbf{261.47} & \textbf{31.52} & \textbf{20.53} & 125.22 & 192.26 & \textbf{82.42} & \textbf{53.68} & 397.28 & 936.46 & \textbf{25.98} & \textbf{11.02} \\
Meanflow (B/4)$^{\star}$ & 1 & 1.0 & \textbf{15.65} & 122.40 & 29.36 & 19.48 & \textbf{53.30} & \textbf{80.34} & 35.94 & 23.84 & \textbf{181.55} & \textbf{412.42} & 10.55 & 4.64 \\
Meanflow (B/4)$^{\star}$ & 25 & 1.0 & 16.28 & 164.63 & 29.91 & 19.78 & 54.43 & 82.31 & 49.24 & 32.56 & 181.98 & 416.10 & 14.73 & 6.44 \\
\midrule

iMF (XL/2) & 1 & 7.0 & 14.36 & 224.85 & 30.20 & 20.15 & 47.55 & 71.27 & 67.90 & 45.31 & 157.45 & 353.68 & 20.51 & 9.13 \\
iMF (XL/2) & 25 & 7.0 & 17.94 & \textbf{258.59} & \textbf{30.50} & \textbf{20.44} & 58.82 & 87.77 & \textbf{78.87} & \textbf{52.86} & 192.85 & 429.40 & \textbf{24.06} & \textbf{10.80} \\
iMF (XL/2)$^{\star}$ & 1 & 6.0 & \textbf{13.91} & 219.30 & 30.14 & 20.13 & \textbf{46.15} & \textbf{69.10} & 66.10 & 44.15 & \textbf{153.12} & \textbf{343.27} & 19.92 & 8.89 \\
iMF (XL/2)$^{\star}$ & 25 & 6.0 & 17.01 & 253.26 & 30.45 & 20.40 & 55.86 & 83.38 & 77.12 & 51.67 & 183.46 & 408.74 & 23.48 & 10.54 \\
\midrule

SoFlow (XL/2)$^{\star}$ & 1 & 2.0 & \textbf{12.80} & 182.80 & 29.96 & 19.95 & \textbf{42.72} & \textbf{64.16} & 54.77 & 36.47 & \textbf{142.60} & \textbf{321.61} & 16.41 & 7.28 \\
SoFlow (XL/2)$^{\star}$ & 25 & 2.0 & 25.76 & \textbf{272.48} & \textbf{30.94} & \textbf{20.70} & 83.26 & 124.44 & \textbf{84.31} & \textbf{56.40} & 269.10 & 601.18 & \textbf{26.08} & \textbf{11.68} \\
\midrule

RAE & 1 & 7.0 & 176.12 & 6.60 & 24.98 & 18.88 & 705.04 & 932.84 & 1.65 & 1.25 & 2822.43 & 4940.88 & 0.41 & 0.23 \\
RAE & 25 & 7.0 & 24.29 & \textbf{282.78} & \textbf{31.30} & \textbf{20.59} & 77.60 & 117.97 & \textbf{88.51} & \textbf{58.22} & 247.94 & 572.95 & \textbf{27.70} & \textbf{11.99} \\
RAE$^{\star}$ & 1 & 1.42 & 305.88 & 3.44 & 24.08 & 18.71 & 1270.27 & 1634.85 & 0.83 & 0.64 & 5275.19 & 8737.83 & 0.20 & 0.12 \\
RAE$^{\star}$ & 25 & 1.42 & \bestblue{11.86} & 199.58 & 30.31 & 20.44 & \bestblue{39.13} & \bestblue{58.02} & 60.49 & 40.79 & \bestblue{129.10} & \bestblue{283.87} & 18.34 & 8.34 \\
\midrule

Scale RAE & 1 & 7.0 & 316.03 & 1.53 & 20.39 & 16.89 & 1549.93 & 1871.11 & 0.31 & 0.26 & 7601.41 & 11078.20 & 0.06 & 0.04 \\
Scale RAE & 25 & 7.0 & 26.99 & 78.80 & \textbf{30.03} & \textbf{21.06} & 89.88 & 128.16 & 23.66 & 16.60 & 299.29 & 608.54 & 7.11 & 3.50 \\
Scale RAE$^{\star}$ & 1 & 1.42 & 316.37 & 1.53 & 20.40 & 16.89 & 1550.83 & 1873.12 & 0.31 & 0.26 & 7602.12 & 11090.11 & 0.06 & 0.04 \\
Scale RAE$^{\star}$ & 25 & 1.42 & \textbf{26.69} & \textbf{80.37} & \textbf{30.01} & \textbf{21.06} & \textbf{88.94} & \textbf{126.73} & \textbf{24.12} & \textbf{16.93} & \textbf{296.36} & \textbf{601.77} & \textbf{7.24} & \textbf{3.56} \\
\midrule

SD3.5 Large & 1 & 7.0 & 182.59 & 2.71 & 24.43 & 18.31 & 747.40 & 997.21 & 0.66 & 0.50 & 3059.36 & 5446.28 & 0.16 & 0.09 \\
SD3.5 Large & 25 & 7.0 & 31.55 & 161.51 & 31.95 & 21.78 & 98.75 & 144.86 & 51.60 & 35.18 & 309.07 & 665.10 & 16.49 & 7.66 \\
SD3.5 Large$^{\star}$ & 1 & 3.5 & 190.13 & 2.66 & 24.52 & 18.60 & 775.41 & 1022.20 & 0.65 & 0.49 & 3162.35 & 5495.72 & 0.16 & 0.09 \\
SD3.5 Large$^{\star}$ & 25 & 3.5 & \textbf{31.69} & \textbf{171.34} & \bestblue{32.10} & \textbf{22.05} & \textbf{98.72} & \textbf{143.72} & \textbf{55.00} & \textbf{37.78} & \textbf{307.55} & \textbf{651.79} & \textbf{17.66} & \textbf{8.33} \\
\midrule

Flux.1 & 1 & 7.0 & 184.82 & 3.95 & 24.51 & 18.90 & 754.06 & 977.88 & 0.97 & 0.75 & 3076.54 & 5173.99 & 0.24 & 0.14 \\
Flux.1 & 25 & 7.0 & 36.24 & 99.05 & 30.39 & 21.79 & 119.25 & 166.31 & 30.10 & 21.58 & 392.40 & 763.26 & 9.15 & 4.70 \\
Flux.1$^{\star}$ & 1 & 3.5 & 198.74 & 3.86 & 25.04 & 19.11 & 793.69 & 1039.98 & 0.97 & 0.74 & 3169.69 & 5442.07 & 0.24 & 0.14 \\
Flux.1$^{\star}$ & 25 & 3.5 & \textbf{34.63} & \textbf{125.78} & \textbf{30.96} & \bestblue{22.13} & \textbf{111.85} & \textbf{156.48} & \textbf{38.94} & \textbf{27.84} & \textbf{361.29} & \textbf{707.11} & \textbf{12.06} & \textbf{6.16} \\
\midrule

SiT ODE & Dynamic & 7.0 & \textbf{12.99} & 196.58 & 29.99 & 20.16 & \textbf{43.31} & \textbf{64.43} & 58.95 & 39.63 & \textbf{144.43} & \textbf{319.62} & 17.68 & 7.99 \\
SiT ODE$^{\star}$ & Dynamic & 1.5 & 38.08 & \bestblue{331.79} & \textbf{31.34} & \textbf{20.84} & 121.51 & 182.73 & \bestblue{103.98} & \bestblue{69.15} & 387.70 & 876.80 & \bestblue{32.59} & \bestblue{14.41} \\

\bottomrule
\end{tabular}
}
\caption{ImageNetV2 benchmark comparing scaled metrics with $\alpha=1$ and $\alpha=2$. $^{\star}$ denotes model-specific tuned CFG. SoFlow prevents custom CFG and SiT prevents explicit step control. Bold marks the best within each model family. Blue marks the best value across all models for each metric.}
\label{tab:imagenetv2_alpha_comparison}
\end{table}

\pagebreak
\subsection{reLAIONet}
\begin{table}[!h]
\centering
\scriptsize
\setlength{\tabcolsep}{0.9pt}
\renewcommand{\arraystretch}{1.08}
\makebox[\linewidth][c]{%
\begin{tabular}{lcccccccccccccc}
\toprule
\textbf{Model} &
\textbf{Steps} & \textbf{CFG} &
\textbf{FID $\downarrow$} & \textbf{IS $\uparrow$} &
\textbf{CLIP $\uparrow$} & \textbf{Pick $\uparrow$} &
\textbf{csFID$_1$ $\downarrow$} & \textbf{psFID$_1$ $\downarrow$} &
\textbf{csIS$_1$ $\uparrow$} & \textbf{psIS$_1$ $\uparrow$} &
\textbf{csFID$_2$ $\downarrow$} & \textbf{psFID$_2$ $\downarrow$} &
\textbf{csIS$_2$ $\uparrow$} & \textbf{psIS$_2$ $\uparrow$} \\
\midrule

Meanflow (B/4) & 1 & 7.0 & 118.35 & 13.37 & 26.65 & 18.40 & 444.09 & 643.21 & 3.56 & 2.46 & 1666.38 & 3495.69 & 0.95 & 0.45 \\
Meanflow (B/4) & 25 & 7.0 & 25.50 & \textbf{245.25} & \textbf{31.71} & \textbf{20.62} & 80.42 & 123.67 & \textbf{77.77} & \textbf{50.57} & 253.60 & 599.74 & \textbf{24.66} & \textbf{10.43} \\
Meanflow (B/4)$^{\star}$ & 1 & 1.0 & 17.85 & 121.92 & 29.56 & 19.53 & 60.39 & 91.40 & 36.04 & 23.81 & 204.28 & 467.99 & 10.65 & 4.65 \\
Meanflow (B/4)$^{\star}$ & 25 & 1.0 & \textbf{13.52} & 158.29 & 30.11 & 19.84 & \textbf{44.90} & \textbf{68.15} & 47.66 & 31.40 & \textbf{149.13} & \textbf{343.47} & 14.35 & 6.23 \\
\midrule

iMF (XL/2) & 1 & 7.0 & 10.82 & 215.89 & 30.31 & 20.21 & 35.70 & 53.54 & 65.44 & 43.63 & 117.78 & 264.91 & 19.83 & 8.82 \\
iMF (XL/2) & 25 & 7.0 & 11.95 & \textbf{247.92} & \textbf{30.63} & \textbf{20.52} & 39.01 & 58.24 & \textbf{75.94} & \textbf{50.87} & 127.37 & 283.80 & \textbf{23.26} & \textbf{10.44} \\
iMF (XL/2)$^{\star}$ & 1 & 6.0 & \textbf{10.64} & 211.18 & 30.27 & 20.18 & \textbf{35.15} & \textbf{52.73} & 63.92 & 42.62 & \textbf{116.12} & \textbf{261.28} & 19.35 & 8.60 \\
iMF (XL/2)$^{\star}$ & 25 & 6.0 & 11.45 & 242.27 & 30.57 & 20.48 & 37.46 & 55.91 & 74.06 & 49.62 & 122.52 & 272.99 & 22.64 & 10.16 \\
\midrule

SoFlow (XL/2)$^{\star}$ & 1 & 2.0 & \textbf{11.13} & 182.70 & 30.15 & 20.00 & \textbf{36.92} & \textbf{55.65} & 55.08 & 36.54 & \textbf{122.44} & \textbf{278.25} & 16.61 & 7.31 \\
SoFlow (XL/2)$^{\star}$ & 25 & 2.0 & 16.87 & \textbf{251.62} & \textbf{31.08} & \textbf{20.77} & 54.28 & 81.22 & \textbf{78.20} & \textbf{52.26} & 174.64 & 391.06 & \textbf{24.31} & \textbf{10.86} \\
\midrule

RAE & 1 & 7.0 & 170.29 & 6.75 & 24.85 & 18.88 & 685.27 & 901.96 & 1.68 & 1.27 & 2757.63 & 4777.33 & 0.42 & 0.24 \\
RAE & 25 & 7.0 & 15.02 & \textbf{259.25} & \textbf{31.51} & \textbf{20.67} & 47.67 & 72.67 & \textbf{81.69} & \textbf{53.59} & 151.28 & 351.55 & \textbf{25.74} & \textbf{11.08} \\
RAE$^{\star}$ & 1 & 1.42 & 296.52 & 3.36 & 24.06 & 18.75 & 1232.42 & 1581.44 & 0.81 & 0.63 & 5122.27 & 8434.35 & 0.20 & 0.12 \\
RAE$^{\star}$ & 25 & 1.42 & \textbf{10.79} & 193.51 & 30.47 & 20.49 & \textbf{35.41} & \textbf{52.66} & 58.96 & 39.65 & \textbf{116.22} & \textbf{257.00} & 17.97 & 8.12 \\
\midrule

Scale RAE & 1 & 7.0 & 319.00 & 1.53 & 20.38 & 16.90 & 1565.26 & 1887.57 & 0.31 & 0.26 & 7680.37 & 11169.08 & 0.06 & 0.04 \\
Scale RAE & 25 & 7.0 & \textbf{22.91} & \textbf{86.23} & \textbf{30.34} & \textbf{21.19} & \textbf{75.51} & \textbf{108.12} & \textbf{26.16} & \textbf{18.27} & \textbf{248.88} & \textbf{510.23} & \textbf{7.94} & \textbf{3.87} \\
Scale RAE$^{\star}$ & 1 & 1.42 & 318.57 & 1.53 & 20.38 & 16.90 & 1563.15 & 1885.03 & 0.31 & 0.26 & 7670.02 & 11154.02 & 0.06 & 0.04 \\
Scale RAE$^{\star}$ & 25 & 1.42 & 22.92 & 85.29 & \textbf{30.34} & \textbf{21.19} & 75.54 & 108.16 & 25.88 & 18.07 & 248.99 & 510.45 & 7.85 & 3.83 \\
\midrule

SiT ODE & Dynamic & 7.0 & 23.07 & \bestblue{309.06} & \textbf{31.56} & \textbf{20.94} & 73.10 & 110.17 & \bestblue{97.54} & \bestblue{64.72} & 231.62 & 526.13 & \bestblue{30.78} & \bestblue{13.55} \\
SiT ODE$^{\star}$ & Dynamic & 1.5 & \bestblue{10.18} & 190.23 & 30.18 & 20.24 & \bestblue{33.73} & \bestblue{50.30} & 57.41 & 38.50 & \bestblue{111.77} & \bestblue{248.50} & 17.33 & 7.79 \\
\midrule

SD3.5 Large & 1 & 7.0 & 178.19 & 2.68 & 24.39 & 18.31 & 730.59 & 973.18 & 0.65 & 0.49 & 2995.43 & 5315.04 & 0.16 & 0.09 \\
SD3.5 Large & 25 & 7.0 & 21.12 & 171.48 & 32.12 & 21.85 & 65.75 & 96.66 & 55.08 & 37.47 & 204.71 & 442.38 & 17.69 & 8.19 \\
SD3.5 Large$^{\star}$ & 1 & 3.5 & 181.90 & 2.64 & 24.52 & 18.61 & 741.84 & 977.43 & 0.65 & 0.49 & 3025.46 & 5252.18 & 0.16 & 0.09 \\
SD3.5 Large$^{\star}$ & 25 & 3.5 & \textbf{20.60} & \textbf{180.44} & \bestblue{32.26} & \textbf{22.13} & \textbf{63.86} & \textbf{93.09} & \textbf{58.21} & \textbf{39.93} & \textbf{197.94} & \textbf{420.63} & \textbf{18.78} & \textbf{8.84} \\
\midrule

Flux.1 & 1 & 7.0 & 182.46 & 4.09 & 24.69 & 18.95 & 739.00 & 962.85 & 1.01 & 0.78 & 2993.13 & 5081.00 & 0.25 & 0.15 \\
Flux.1 & 25 & 7.0 & 29.29 & 108.79 & 30.73 & 21.90 & 95.31 & 133.74 & 33.43 & 23.83 & 310.17 & 610.71 & 10.27 & 5.22 \\
Flux.1$^{\star}$ & 1 & 3.5 & 200.32 & 3.99 & 25.17 & 19.15 & 795.87 & 1046.06 & 1.00 & 0.76 & 3161.97 & 5462.44 & 0.25 & 0.15 \\
Flux.1$^{\star}$ & 25 & 3.5 & \textbf{26.45} & \textbf{131.72} & \textbf{31.24} & \bestblue{22.24} & \textbf{84.67} & \textbf{118.93} & \textbf{41.15} & \textbf{29.29} & \textbf{271.02} & \textbf{534.76} & \textbf{12.86} & \textbf{6.51} \\

\bottomrule
\end{tabular}
}
\caption{reLAIONet benchmark comparing scaled metrics with $\alpha=1$ and $\alpha=2$. $^{\star}$ denotes model-specific tuned CFG. SoFlow prevents custom CFG and SiT prevents explicit step control. Bold marks the best within each model family. Blue marks the best value across all models for each metric.}
\label{tab:relaionet_alpha_comparison}
\end{table}

\section{Menaflow and iMF Ablations Raw Data}
\subsection{Meanflow Ablation Raw Results}
\begin{table}[H]
\centering
\footnotesize
\setlength{\tabcolsep}{2.5pt}
\label{tab:meanflow}
\begin{tabular}{rrrrrrrrrr}
\toprule
CFG Scale & Steps & FID$\downarrow$ & IS Mean$\uparrow$ & CLIP Score$\uparrow$ & PickScore$\uparrow$ & csFID$\downarrow$ & psFID$\downarrow$ & csIS$\uparrow$ & psIS$\uparrow$ \\
\midrule
\multirow{6}{*}{1}
 & 1  & 10.01 & 150.48 & 29.36 & 19.49 & 116.12 & 263.52 & 12.97 & 5.72 \\
 & 5  &  7.74 & 198.98 & 29.84 & 19.75 &  86.92 & 198.43 & 17.72 & 7.76 \\
 & 10 &  7.69 & 202.32 & 29.88 & 19.77 &  86.13 & 196.75 & 18.06 & 7.91 \\
 & 15 &  7.70 & 203.87 & 29.89 & 19.78 &  86.19 & 196.81 & 18.21 & 7.98 \\
 & 20 &  7.72 & 204.65 & 29.90 & 19.78 &  86.35 & 197.32 & 18.30 & 8.01 \\
 & 25 &  7.74 & 205.41 & 29.90 & 19.79 &  86.58 & 197.63 & 18.36 & 8.04 \\
\midrule
\multirow{6}{*}{3}
 & 1  & 56.77 &  35.90 & 28.49 & 18.95 &  699.41 & 1580.89 &  2.91 &  1.29 \\
 & 5  & 16.93 & 258.28 & 30.85 & 20.08 &  177.89 &  419.88 & 24.58 & 10.41 \\
 & 10 & 20.76 & 338.93 & 31.31 & 20.39 &  211.77 &  499.34 & 33.23 & 14.09 \\
 & 15 & 22.10 & 355.46 & 31.41 & 20.47 &  224.00 &  527.42 & 35.07 & 14.89 \\
 & 20 & 22.60 & 360.88 & 31.44 & 20.50 &  228.64 &  537.78 & 35.67 & 15.17 \\
 & 25 & 22.81 & 364.07 & 31.46 & 20.51 &  230.47 &  542.24 & 36.03 & 15.31 \\
\midrule
\multirow{6}{*}{6}
 & 1  & 103.76 &  16.41 & 26.83 & 18.47 & 1441.41 & 3041.56 &  1.18 &  0.56 \\
 & 5  &  30.76 &  86.62 & 29.00 & 19.33 &  365.76 &  823.23 &  7.28 &  3.24 \\
 & 10 &  20.97 & 237.92 & 30.83 & 20.13 &  220.62 &  517.50 & 22.61 &  9.64 \\
 & 15 &  23.17 & 318.27 & 31.34 & 20.42 &  235.90 &  555.67 & 31.26 & 13.27 \\
 & 20 &  24.71 & 345.52 & 31.50 & 20.52 &  249.03 &  586.84 & 34.28 & 14.55 \\
 & 25 &  25.29 & 356.21 & 31.54 & 20.56 &  254.23 &  598.28 & 35.43 & 15.06 \\
\midrule
\multirow{6}{*}{7}
 & 1  & 113.34 &  13.94 & 26.61 & 18.41 & 1600.64 & 3344.07 &  0.99 &  0.47 \\
 & 5  &  38.70 &  60.69 & 28.34 & 19.11 &  481.85 & 1059.72 &  4.87 &  2.22 \\
 & 10 &  21.47 & 195.07 & 30.50 & 19.97 &  230.80 &  538.36 & 18.15 &  7.78 \\
 & 15 &  22.42 & 289.36 & 31.20 & 20.33 &  230.32 &  542.45 & 28.17 & 11.96 \\
 & 20 &  24.15 & 330.08 & 31.44 & 20.47 &  244.32 &  576.34 & 32.63 & 13.83 \\
 & 25 &  25.12 & 345.75 & 31.51 & 20.52 &  253.00 &  596.57 & 34.33 & 14.56 \\
\midrule
\multirow{6}{*}{9}
 & 1  & 126.07 &  10.88 & 26.36 & 18.34 & 1814.35 & 3748.12 &  0.76 &  0.37 \\
 & 5  &  54.77 &  34.17 & 27.21 & 18.78 &  739.75 & 1552.93 &  2.53 &  1.21 \\
 & 10 &  25.52 & 123.12 & 29.73 & 19.62 &  288.73 &  662.95 & 10.88 &  4.74 \\
 & 15 &  21.37 & 221.82 & 30.78 & 20.10 &  225.56 &  528.95 & 21.02 &  8.96 \\
 & 20 &  22.55 & 281.75 & 31.20 & 20.32 &  231.65 &  546.13 & 27.43 & 11.63 \\
 & 25 &  23.83 & 311.44 & 31.39 & 20.42 &  241.85 &  571.50 & 30.69 & 12.99 \\
\midrule
\multirow{6}{*}{12}
 & 1  & 136.16 &   8.73 & 26.17 & 18.30 & 1988.12 & 4065.81 &  0.60 &  0.29 \\
 & 5  &  75.73 &  18.99 & 25.99 & 18.48 & 1121.13 & 2217.50 &  1.28 &  0.65 \\
 & 10 &  37.41 &  62.59 & 28.52 & 19.17 &  459.93 & 1017.99 &  5.09 &  2.30 \\
 & 15 &  24.68 & 132.84 & 29.89 & 19.68 &  276.24 &  637.23 & 11.87 &  5.14 \\
 & 20 &  21.87 & 195.37 & 30.62 & 20.00 &  233.26 &  546.75 & 18.32 &  7.81 \\
 & 25 &  22.08 & 240.84 & 30.99 & 20.17 &  229.91 &  542.73 & 23.13 &  9.80 \\
\midrule
\multirow{6}{*}{15}
 & 1  & 141.47 &   7.70 & 26.06 & 18.28 & 2083.13 & 4233.62 &  0.52 &  0.26 \\
 & 5  &  91.98 &  13.30 & 25.16 & 18.31 & 1453.02 & 2743.57 &  0.84 &  0.45 \\
 & 10 &  52.25 &  35.39 & 27.46 & 18.83 &  692.92 & 1473.62 &  2.67 &  1.25 \\
 & 15 &  33.81 &  76.44 & 28.92 & 19.30 &  404.25 &  907.68 &  6.39 &  2.85 \\
 & 20 &  25.88 & 125.57 & 29.85 & 19.64 &  290.45 &  670.94 & 11.19 &  4.84 \\
 & 25 &  23.06 & 170.15 & 30.40 & 19.87 &  249.52 &  584.07 & 15.72 &  6.72 \\
\bottomrule
\end{tabular}
\caption{Ablations for Meanflow across step sizes of 1, 5, 10, 15, 20, 25 and CFG scales of 1, 3, 6, 7, 9, 12, 15 on ImageNet, including CLIP and Pick Scaled FID/IS}
\end{table}

\subsection{Improved Meanflow Ablation Raw Results}

\begin{table}[H]
\centering
\footnotesize
\setlength{\tabcolsep}{2.5pt}
\label{tab:imeanflow}
\begin{tabular}{rrrrrrrrrr}
\toprule
CFG Scale & Steps & FID$\downarrow$ & IS Mean$\uparrow$ & CLIP Score$\uparrow$ & PickScore$\uparrow$ & csFID$\downarrow$ & psFID$\downarrow$ & csIS$\uparrow$ & psIS$\uparrow$ \\
\midrule
\multirow{6}{*}{1}
 & 1  & 9.93 & 123.37 & 21.34 & 18.27 & 218.05 & 297.49 & 5.62 & 4.12 \\
 & 5  & 6.08 & 141.77 & 28.89 & 19.72 & 72.85 & 156.35 & 11.83 & 5.51 \\
 & 10 & 5.89 & 142.62 & 28.90 & 19.73 & 70.52 & 151.31 & 11.91 & 5.55 \\
 & 15 & 5.84 & 143.20 & 28.92 & 19.73 & 69.83 & 150.02 & 11.98 & 5.57 \\
 & 20 & 5.82 & 143.56 & 28.93 & 19.74 & 69.54 & 149.36 & 12.02 & 5.59 \\
 & 25 & 5.78 & 144.24 & 28.93 & 19.74 & 69.06 & 148.33 & 12.07 & 5.62 \\
\midrule
\multirow{6}{*}{3}
 & 1  & 3.37 & 229.94 & 29.75 & 19.95 & 38.08 & 84.67 & 20.35 & 9.15 \\
 & 5  & 3.12 & 261.17 & 29.96 & 20.14 & 34.76 & 76.92 & 23.44 & 10.59 \\
 & 10 & 3.19 & 268.08 & 30.00 & 20.16 & 35.44 & 78.49 & 24.13 & 10.90 \\
 & 15 & 3.22 & 269.46 & 30.02 & 20.17 & 35.73 & 79.15 & 24.28 & 10.96 \\
 & 20 & 3.23 & 269.05 & 30.02 & 20.17 & 35.84 & 79.39 & 24.25 & 10.95 \\
 & 25 & 3.24 & 268.03 & 30.03 & 20.18 & 35.93 & 79.56 & 24.17 & 10.92 \\
\midrule
\multirow{6}{*}{6}
 & 1  & 3.85 & 285.96 & 30.15 & 20.13 & 42.35 & 95.01 & 25.99 & 11.59 \\
 & 5  & 5.05 & 323.10 & 30.36 & 20.35 & 54.79 & 121.94 & 29.78 & 13.38 \\
 & 10 & 5.41 & 333.12 & 30.40 & 20.39 & 58.54 & 130.13 & 30.79 & 13.85 \\
 & 15 & 5.54 & 335.80 & 30.41 & 20.40 & 59.91 & 133.12 & 31.05 & 13.97 \\
 & 20 & 5.57 & 336.21 & 30.42 & 20.41 & 60.19 & 133.71 & 31.11 & 14.01 \\
 & 25 & 5.62 & 336.30 & 30.43 & 20.41 & 60.69 & 134.91 & 31.14 & 14.01 \\
\midrule
\multirow{6}{*}{7}
 & 1  & 4.07 & 293.12 & 30.20 & 20.15 & 44.63 & 100.24 & 26.73 & 11.90 \\
 & 5  & 5.57 & 334.34 & 30.42 & 20.39 & 60.19 & 133.97 & 30.94 & 13.90 \\
 & 10 & 6.00 & 342.93 & 30.46 & 20.43 & 64.67 & 143.75 & 31.82 & 14.31 \\
 & 15 & 6.11 & 345.11 & 30.47 & 20.44 & 65.81 & 146.24 & 32.04 & 14.42 \\
 & 20 & 6.17 & 346.91 & 30.48 & 20.45 & 66.41 & 147.54 & 32.23 & 14.51 \\
 & 25 & 6.20 & 348.08 & 30.48 & 20.45 & 66.74 & 148.25 & 32.34 & 14.56 \\
\midrule
\multirow{6}{*}{9}
 & 1  & 4.29 & 298.69 & 30.25 & 20.17 & 46.88 & 105.45 & 27.33 & 12.15 \\
 & 5  & 6.16 & 343.75 & 30.48 & 20.43 & 66.31 & 147.59 & 31.94 & 14.35 \\
 & 10 & 6.68 & 352.78 & 30.52 & 20.47 & 71.71 & 159.42 & 32.86 & 14.78 \\
 & 15 & 6.79 & 354.83 & 30.54 & 20.49 & 72.80 & 161.73 & 33.09 & 14.90 \\
 & 20 & 6.88 & 357.28 & 30.55 & 20.50 & 73.72 & 163.71 & 33.35 & 15.01 \\
 & 25 & 6.94 & 357.07 & 30.55 & 20.50 & 74.36 & 165.14 & 33.33 & 15.01 \\
\midrule
\multirow{6}{*}{12}
 & 1  & 4.04 & 291.44 & 30.19 & 20.14 & 44.33 & 99.60 & 26.56 & 11.82 \\
 & 5  & 5.79 & 335.62 & 30.44 & 20.41 & 62.49 & 138.99 & 31.10 & 13.98 \\
 & 10 & 6.24 & 346.55 & 30.48 & 20.45 & 67.17 & 149.21 & 32.20 & 14.49 \\
 & 15 & 6.38 & 348.87 & 30.50 & 20.46 & 68.58 & 152.41 & 32.45 & 14.60 \\
 & 20 & 6.53 & 349.83 & 30.51 & 20.47 & 70.15 & 155.84 & 32.56 & 14.66 \\
 & 25 & 6.60 & 351.48 & 30.52 & 20.48 & 70.86 & 157.36 & 32.74 & 14.74 \\
\midrule
\multirow{6}{*}{15}
 & 1  & 3.68 & 278.09 & 30.08 & 20.08 & 40.67 & 91.27 & 25.16 & 11.21 \\
 & 5  & 5.32 & 327.97 & 30.39 & 20.37 & 57.60 & 128.21 & 30.29 & 13.61 \\
 & 10 & 5.70 & 336.33 & 30.43 & 20.40 & 61.56 & 136.97 & 31.14 & 14.00 \\
 & 15 & 5.90 & 340.55 & 30.45 & 20.42 & 63.63 & 141.49 & 31.58 & 14.20 \\
 & 20 & 6.01 & 340.46 & 30.46 & 20.43 & 64.78 & 143.99 & 31.59 & 14.21 \\
 & 25 & 6.10 & 342.97 & 30.47 & 20.44 & 65.70 & 146.01 & 31.84 & 14.33 \\
\bottomrule
\end{tabular}
\caption{Ablations for Improved Meanflow across step sizes of 1, 5, 10, 15, 20, 25 and CFG scales of 1, 3, 6, 7, 9, 12, 15 on ImageNet, including CLIP and Pick Scaled FID/IS}
\end{table}
\pagebreak

\section{Heatmap of Ablations for iMF}
\begin{figure}[H]
    \centering
    \makebox[\linewidth][c]{%
    \includegraphics[width=0.8\textwidth]{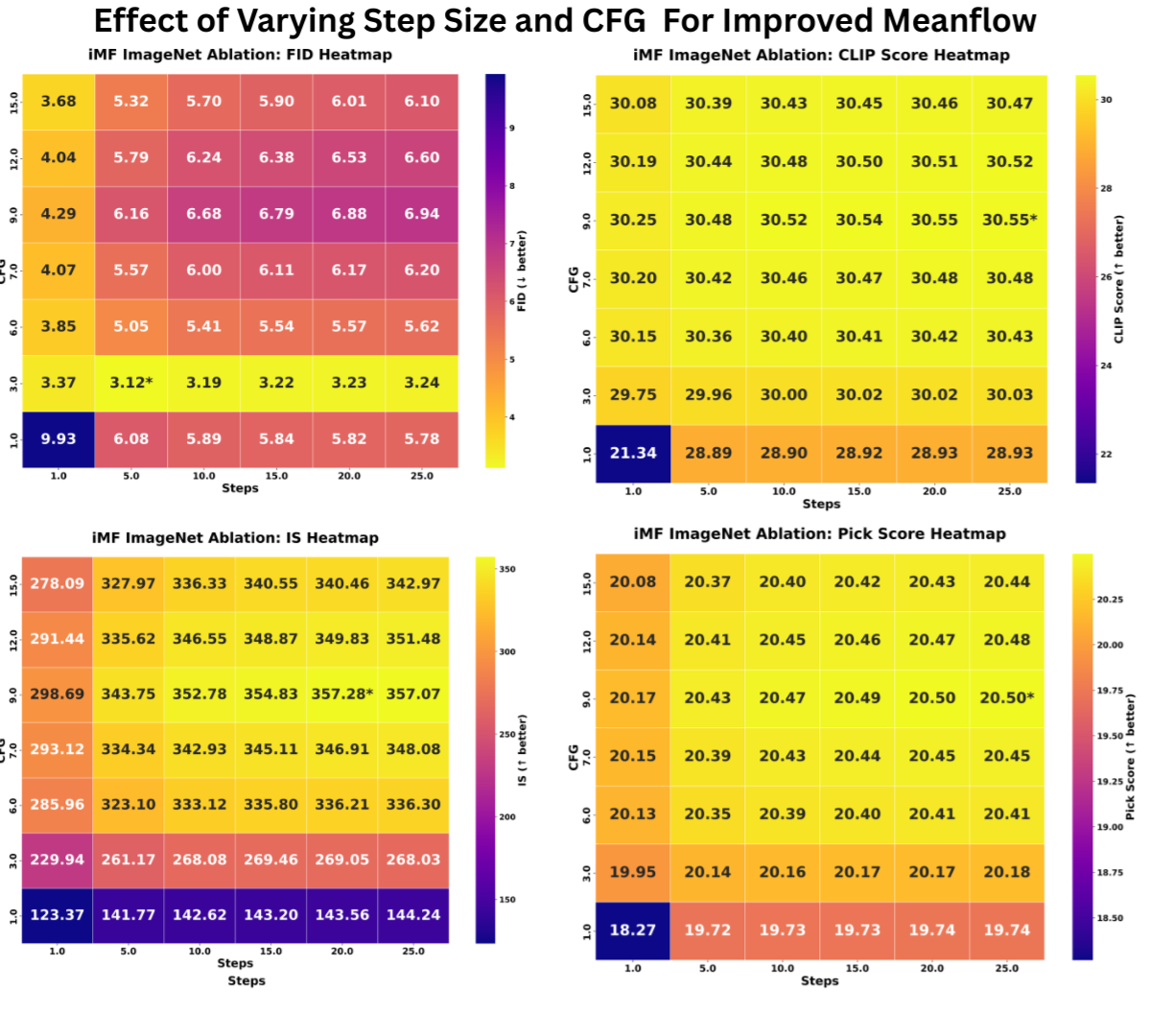}
    }
    \caption{Effect of step size and CFG on Improved MeanFlow performance. Heatmaps show FID, CLIP Score, Inception Score, and PickScore across different step sizes and CFG values}
\end{figure}

\section{reLAIONet Class Distribution}
\begin{figure}[H]
    \centering
    \makebox[\linewidth][c]{%
    \includegraphics[width=1.2\textwidth]{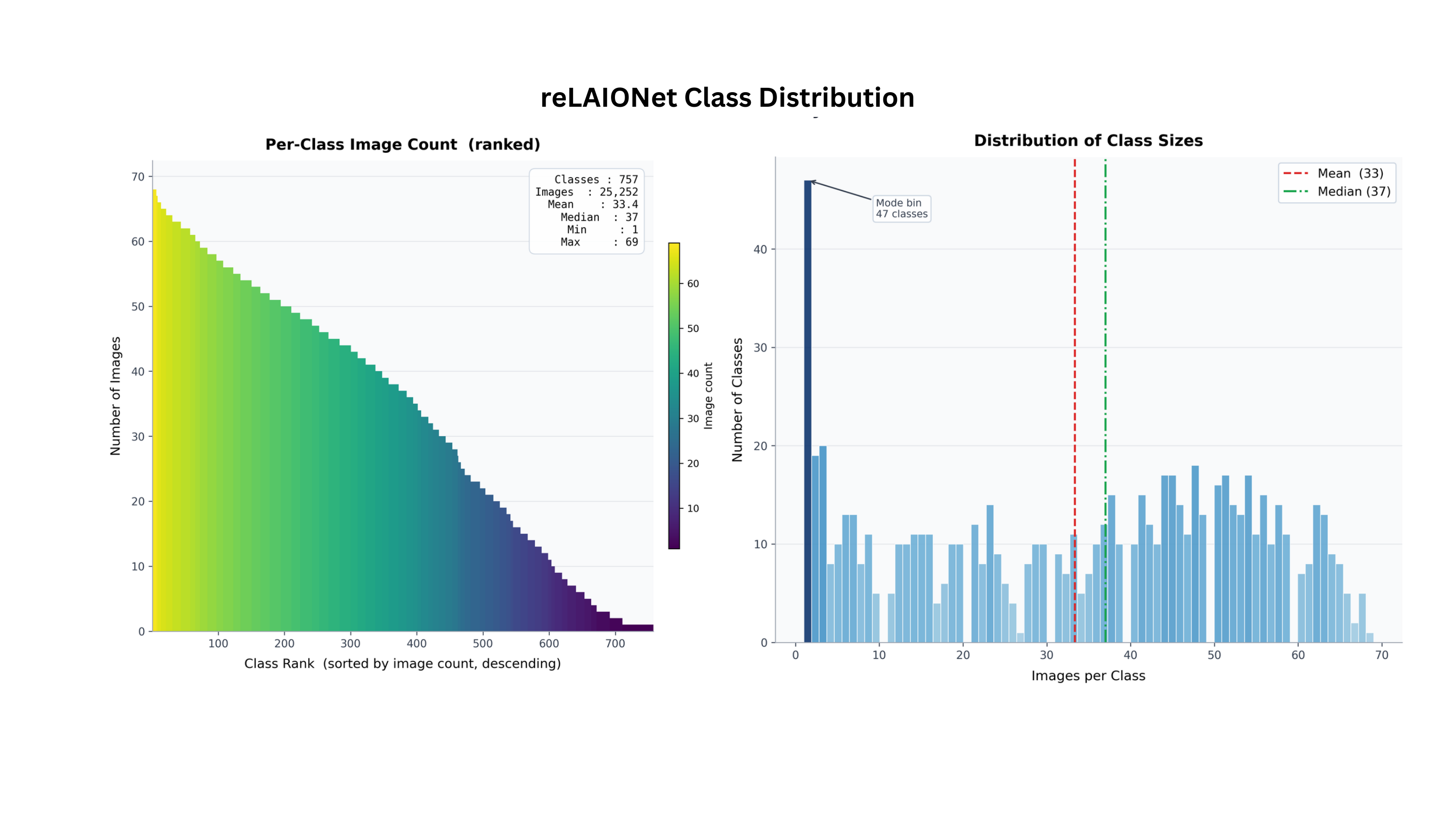}
    }
    \caption{\textbf{reLAIONet class distribution.}
Left: Ranked per-class image counts after manual filtering, showing that reLAIONet covers 757 ImageNet classes with 25,252 total images (x-axis is a reindexing and does not reflect class labels). Right: Histogram of class sizes, with mean 33 and median 37 images per class. Most classes contain multiple validated samples, while long-tail classes retain fewer images after proofreading.}
\end{figure}

\pagebreak
\section{ReLAIONet Setup and Construction} \label{sec:relaionet_construction}
\subsection{Overview and Hyperparameters}

reLAIONet is constructed by streaming all shards of the reLAION-400M
metadata, filtering candidates via lexical and semantic criteria, and
downloading the highest-scoring images per class.
The resulting dataset comprises \textbf{25,251 images} spanning
\textbf{757 ImageNet classes}, with up to $K$ images per class ranked
by CLIP cosine similarity to their synset text.

\begin{table}[h]
\centering
\caption{LAIONet construction hyperparameters.}
\label{tab:hparams}
\begin{tabular}{lrl}
\toprule
Parameter & Value & Description \\
\midrule
$\tau$        & $0.82$      & CLIP cosine similarity threshold \\
$N$           & $1000$        & Top-$N$ classes selected \\
$K$           & $70$   & Max images per class \\
$B$           & $2{,}048$   & CLIP encoding batch size \\
CLIP model    & ViT-B/32    & OpenAI pre-trained weights \\
Lemma filter  & unique only & Ambiguous lemmas excluded \\
\bottomrule
\end{tabular}
\end{table}

\subsection{Lemmas}

Each of the 1{,}000 ILSVRC-2012 synsets $S_k = (w_k, \mathcal{L}_k, d_k, k)$
provides a set of WordNet lemma strings $\mathcal{L}_k$.
Lemmas are normalised by lowercasing and replacing underscores with
spaces, yielding $\hat{\mathcal{L}}_k$.
The unique lemma index $\phi : \hat\ell \mapsto w_k$ retains only
lemmas that appear in exactly one synset. Any lemma shared across
multiple synsets (e.g., \texttt{crane}, which maps to both a bird and
a construction machine) is excluded to prevent ambiguous assignments.

\begin{lemma}[Uniqueness Exclusion]
A caption containing an ambiguous lemma $\hat\ell$ with
$|\{k : \hat\ell \in \hat{\mathcal{L}}_k\}| > 1$ cannot be
unambiguously assigned to a single class and is discarded.
\end{lemma}

\begin{lemma}[Multi-label Rejection]
A caption matching two or more distinct WNIDs via $\phi$ is assigned
the sentinel \texttt{\_\_multi\_\_} and removed from the candidate pool.
\end{lemma}

\begin{lemma}[Threshold Monotonicity]
Increasing the CLIP threshold $\tau$ weakly decreases the candidate
set size for every class, and weakly increases the average semantic
fidelity of retained captions.
\end{lemma}

For CLIP scoring, each synset is represented by the text
$\tau_k = \texttt{``}\langle name \rangle\texttt{ which is }
\langle definition \rangle\texttt{''}$,
encoded with CLIP ViT-B/32 and $\ell_2$-normalised so that cosine
similarity equals the dot product.

\subsection{Class Removal Criteria}

A class or individual image is excluded if any of the following
criteria is satisfied.

\begin{enumerate}
\item \textbf{Synset ID mismatch.}
The assigned WNID does not correspond to the WordNet synset that
describes the depicted object.

\item \textbf{ILSVRC-2012 ID mismatch.}
The class is not among the 1{,}000 official ILSVRC-2012 WNIDs, or the
class index is inconsistent with the canonical ILSVRC devkit.

\item \textbf{Text-dominant imagery.}
The majority of the image area consists of text, diagrams, or
infographics rather than the target object, providing little visual
signal for recognition.

\item \textbf{Insensitive or joke content.}
The image or caption constitutes satirical or joke content that is
broadly considered insensitive or offensive.

\item \textbf{NSFW content.}
The image contains sexually explicit material, graphic violence, or
other content unsuitable for general audiences, as flagged by the
reLAION-400M NSFW metadata field or manual review.

\end{enumerate}

We perform 2 rounds of review split into batches. The data was was cleaned by 12 people and then the cleaned batches were distributed to a new annotator to reduce errors.

\section{Evaluation Hyperparameters For Each Model}
\begin{table}[H]
\centering
\label{tab:scale-rae}
\begin{tabular}{ll}
\toprule
\textbf{Hyperparameter} & \textbf{Value} \\
\midrule
Model                   & \texttt{nyu-visionx/Scale-RAE-Qwen1.5B\_DiT2.4B} \\
Decoder encoder         & \texttt{google/siglip2-so400m-patch14-224} \\
Max new tokens          & 512 \\
Conditioning            & Text prompt: ``Can you generate a photo of a \{class\}?'' \\
Image size (metrics)    & $256\times256$ \\
Resize filter           & LANCZOS \\
Seed                    & 42 \\
\bottomrule
\end{tabular}
\caption{Scale-RAE Evaluation Setup}
\end{table}

\begin{table}[H]
\centering

\label{tab:sit}
\begin{tabular}{ll}
\toprule
\textbf{Hyperparameter} & \textbf{Value} \\
\midrule
Model          & SiT-XL/2 \\
VAE                     & \texttt{stabilityai/sd-vae-ft-mse} \\
Image size              & $256\times256$ \\
Latent size             & $32\times32$ \\
Sampler mode            & ODE (Euler) \\
Generation batch size   & 16 \\
Real image batch size   & 128 \\
Metric batch size       & 64 \\
Seed                    & 42 \\
\bottomrule
\end{tabular}
\caption{SiT Evaluation Setup}
\end{table}

\begin{table}[H]
\centering
\label{tab:sd35}
\begin{tabular}{ll}
\toprule
\textbf{Hyperparameter} & \textbf{Value} \\
\midrule
Model                   & \texttt{stabilityai/stable-diffusion-3.5-large} \\
Conditioning            & Text prompt: ``a photo of a \{class\}'' \\
Metric resolution       & $256\times256$ (LANCZOS resize) \\
Generation batch size   & 4 \\
Seed                    & 42 \\
\bottomrule
\end{tabular}
\caption{SD 3.5 Large Evaluation Setup}
\end{table}

\begin{table}[H]
\centering
\label{tab:flux}
\begin{tabular}{ll}
\toprule
\textbf{Hyperparameter} & \textbf{Value} \\
\midrule
Model                   & \texttt{black-forest-labs/FLUX.1-dev} \\
Conditioning            & Text prompt: ``a photo of a \{class\}'' \\
Metric resolution       & $256\times256$ (LANCZOS resize) \\
Generation batch size   & 1 \\
Seed                    & 42 \\
\bottomrule
\end{tabular}
\caption{FLUX.1 [dev] Evaluation Setup}
\end{table}

\begin{table}[H]
\centering
\label{tab:meanflow_setup}
\begin{tabular}{ll}
\toprule
\textbf{Hyperparameter} & \textbf{Value} \\
\midrule
Model variant           & DiT\_B\_4 \\
VAE                     & \texttt{stabilityai/sd-vae-ft-mse} \\
Image size (metrics)    & $256\times256$ \\
Latent size             & $32\times32\times4$ \\
Sampler                 & ODE (Euler, \texttt{fori\_loop}) \\
Per-device batch size   & 1 \\
Seed                    & 42 \\
\bottomrule
\end{tabular}
\caption{MeanFlow Evaluation Setup}
\end{table}

\begin{table}[H]
\centering
\label{tab:imeanflow}
\begin{tabular}{ll}
\toprule
\textbf{Hyperparameter} & \textbf{Value} \\
\midrule
Model variant           & MiT\_XL\_2 \\
VAE                     & \texttt{stabilityai/sd-vae-ft-mse} \\
Image size (metrics)    & $256\times256$ \\
Latent size             & $32\times32\times4$ \\
Sampler                 & ODE with CFG interval \\
Per-device batch size   & 1 \\
Seed                    & 42 \\
\bottomrule
\end{tabular}
\caption{Improved MeanFlow Evaluation Setup}
\end{table}

\begin{table}[H]
\centering
\label{tab:soflow}
\begin{tabular}{ll}
\toprule
\textbf{Hyperparameter} & \textbf{Value} \\
\midrule
Model variant           & DiT-XL-2-cond \\
VAE                     & \texttt{stabilityai/sd-vae-ft-mse} \\
Image size (metrics)    & $256\times256$ \\
Latent size             & $32\times32\times4$ \\
Per-device batch size   & 32 \\
Seed                    & 42 \\
\bottomrule
\end{tabular}
\caption{SoFlow Evaluation Setup}
\end{table}

\begin{table}[H]
\centering
\label{tab:rae}
\begin{tabular}{ll}
\toprule
\textbf{Hyperparameter} & \textbf{Value} \\
\midrule
Architecture            & Two-stage: Stage 1 RAE decoder + Stage 2 diffusion \\
Latent size             & $(768,16,16)$ (default) \\
Image size (metrics)    & $256\times256$ \\
Per-device batch size   & 25 \\
Seed                    & 42 \\
\bottomrule
\end{tabular}
\caption{RAE Evaluation Setup}
\end{table}

\end{document}